\documentclass[a4paper,fleqn]{cas-dc}

\usepackage[numbers,sort&compress]{natbib}
\usepackage{graphicx}
\usepackage{amsmath,amssymb}
\usepackage{booktabs}
\usepackage{float}
\usepackage{placeins}
\usepackage{ragged2e}
\usepackage{etoolbox}
\usepackage{xspace}

\def\tsc#1{\csdef{#1}{\textsc{\lowercase{#1}}\xspace}}
\tsc{WGM}
\tsc{QE}

\makeatletter
\renewcommand{\fnum@figure}{Fig.\ \thefigure}
\makeatother

\ExplSyntaxOn
\cs_set:Npn \__make_fig_caption:nn #1#2
{
  \l_fig_align_tl
  \skip_vertical:N \l_fig_abovecap_skip
  \setbox\cascaptionbox=\hbox{%
     \rmfamily\normalsize\textbf{\color{scolor}#1.}~#2}
  \ifdim\the\wd\cascaptionbox<\dim_use:N \l_fig_width_dim\relax
    \parbox{ \l_fig_width_dim }
      {\unskip\ignorespaces\hfil\rmfamily\normalsize
       \textbf{\color{scolor}#1.}~#2\hfil\par }
  \else
    \parbox{ \l_fig_width_dim }
      {\rightskip=0pt\unskip\ignorespaces\rmfamily
       \normalsize\textbf{\color{scolor}#1.}~#2\par }
  \fi
  \skip_vertical:N \l_fig_belowcap_skip
}
\cs_set:Npn \__make_tbl_caption:nn #1#2
{
  \l_tbl_align_tl
  \skip_vertical:N \l_tbl_abovecap_skip
  {\parbox{ \dimexpr(\l_tbl_width_dim)}
    {\rightskip=0pt\rmfamily\normalsize\textbf{\color{scolor}#1}\par#2\par\vskip4pt }}
  \skip_vertical:N \l_tbl_belowcap_skip
}
\ExplSyntaxOff

\AtBeginEnvironment{thebibliography}{\RaggedRight}

\begin{document}
\let\WriteBookmarks\relax
\def\floatpagepagefraction{.8}
\def\textpagefraction{.08}
\def\dblfloatpagefraction{.8}
\renewcommand{\topfraction}{0.9}
\renewcommand{\bottomfraction}{0.8}
\renewcommand{\floatpagefraction}{0.75}
\setlength{\textfloatsep}{10pt plus 2pt minus 2pt}
\setlength{\floatsep}{8pt plus 2pt minus 2pt}
\setlength{\intextsep}{8pt plus 2pt minus 2pt}
\setlength{\dbltextfloatsep}{10pt plus 2pt minus 2pt}

\newcommand{\safeincludegraphics}[2][]{%
  \IfFileExists{#2}{\includegraphics[#1]{#2}}{%
    \fbox{\parbox[c][0.25\textheight][c]{0.95\linewidth}{Missing file: \texttt{#2}}}%
  }%
}

\newcommand{\ModuleFloatBarrier}{\FloatBarrier}

\shorttitle{AGE-Net: Spectral--Spatial Fusion and Anatomical Graph Reasoning}

\shortauthors{Xiaoyang Li et al.}

\title[mode=title]{AGE-Net: Spectral--Spatial Fusion and Anatomical Graph Reasoning with Evidential Ordinal Regression for Knee Osteoarthritis Grading}

\author[a]{Xiaoyang Li}[orcid=0009-0002-4863-5761]
\fnmark[1]

\author[a]{Runni Zhou}[orcid=0009-0006-8365-4852]
\fnmark[1]

\author[a]{Xinghao Yan}
\author[a]{Liehao Yan}
\author[a]{Zhaochen Li}
\author[a]{Chenjie Zhu}
\author[a]{Rongrong Fu}

\author[a]{Yuan Chai}[orcid=0000-0001-6977-6155]
\cormark[1]

\affiliation[a]{organization={College of Medicine and Biological Information Engineering, Northeastern University},
               city={Shenyang},
               postcode={110016},
               country={China}}

\cortext[1]{Corresponding author}
\fntext[1]{These authors contributed equally to this work.}

\begin{abstract}
Accurate Kellgren--Lawrence (KL) grading of knee osteoarthritis from plain radiographs remains challenging because of subtle early degenerative changes, long-range anatomical dependencies, and ambiguity near adjacent-grade boundaries. To address these issues, we propose AGE-Net, an anatomy-aware and ordinally constrained deep learning framework for automated KL grading. Built upon a ConvNeXt-Base backbone, AGE-Net integrates three complementary components: Spectral--Spatial Fusion (SSF) for enhancing diagnostically relevant frequency-sensitive textures, a macro--micro Anatomical Graph Reasoning (AGR) module for modeling non-local anatomical relations, and a Pathology-Aware Differential Refinement (PA-DFR) module for emphasizing pathology-relevant boundary cues while suppressing non-informative edges. To better align prediction with the ordered structure of disease severity, we further formulate the task as evidential ordinal regression by combining a Normal--Inverse-Gamma (NIG) evidential head with a pairwise ordinal ranking constraint. We evaluate AGE-Net on a large public knee radiograph cohort and further assess external transferability on NHANES III. Across three random initialization seeds, AGE-Net achieves a quadratic weighted kappa (QWK) of 0.9017$\pm$0.0045 and a mean squared error (MSE) of 0.2349$\pm$0.0028 on the internal test set. External evaluation yields a QWK of 0.8640$\pm$0.0025 and an MSE of 0.3413$\pm$0.0047. Ablation studies support the complementary contribution of the proposed components, while qualitative analyses suggest that model attention is concentrated around clinically relevant joint structures. In addition, the estimated uncertainty is associated with prediction difficulty and enables selective prediction analysis. These findings support AGE-Net as an effective framework for automated KOA severity assessment, while further validation is warranted to establish robustness across broader clinical settings.
\end{abstract}


\maketitle

\section{Introduction}

Knee osteoarthritis (OA) represents one of the most pervasive degenerative musculoskeletal pathologies, precipitating a profound global health burden characterized by chronic nociception, progressive mobility impairment, and substantially diminished quality of life. The radiographic Kellgren--Lawrence (KL) grading system\cite{kellgren1957kl} persists as the paramount clinical gold standard for stratifying OA severity, delineating progression from Grade 0 (unremarkable) to Grade 4 (severe morphological degradation)\cite{peterfy2004worms}. The automation of KL grading via plain radiographs harbors immense potential to alleviate clinical workloads, mitigate inter-observer variability, and expedite population-level diagnostic screening.

Notwithstanding progressive advancements in deep learning applied to musculoskeletal radiography Notwithstanding progressive advancements in deep learning applied to musculoskeletal radiography\cite{tiulpin2018diagnosis}, the formulation of highly reliable KL grading algorithms remains a formidable endeavor owing to three persistent methodological bottlenecks. Primarily, transitional KL grades are frequently dictated by inconspicuous and localized anatomical aberrations—such as nascent osteophyte formation or marginal joint-space narrowing—which are deeply embedded within low-contrast osteochondral contexts. Secondly, definitive OA severity evaluation is contingent upon overarching anatomical coherence; medial and lateral compartmental configurations, tibiofemoral alignment, and spatially distributed textural cues synergistically constitute the overall diagnostic grade, a complexity inherently under-modeled by strictly localized operational receptive fields. Finally, clinical KL labels are intrinsically ordinal and chronically afflicted by boundary ambiguity. Borderline presentations are ubiquitous, precipitating non-negligible label noise even amongst seasoned subspecialty radiologists. Ergo, a clinically viable computational system must inexorably (i) respect the monotonic trajectory of disease progression across grades and (ii) yield calibrated uncertainty distributions that faithfully reflect underlying diagnostic ambiguity and potential error.

Although modern convolutional neural networks (CNNs), especially ConvNeXt\cite{liu2022convnext}, provide strong hierarchical feature representations, conventional end-to-end classification pipelines do not fully address two key challenges in KL grading: anatomy-aligned long-range dependency modeling and uncertainty-aware ordinal prediction. To this end, we propose AGE-Net, a ConvNeXt-based framework composed of four complementary modules: Spectral--Spatial Fusion (SSF) for frequency-aware enhancement of subtle osteophytic textures, an Anatomical Graph Reasoner (AGR) for long-range anatomical interaction modeling, a Pathology-Aware Differential Refinement (PA-DFR) module for boundary-sensitive degenerative cue refinement, and a Continuous Ordinal Evidential Head (COE-Head) that combines Normal--Inverse-Gamma (NIG) evidential regression with a pairwise ordinal ranking constraint for severity-aware prediction and uncertainty estimation.

\textbf{Contributions.} The main contributions of this work are summarized as follows:
\begin{itemize}
    \item We present AGE-Net, an anatomy-aware framework for automated KL grading that integrates Spectral--Spatial Fusion (SSF), Anatomical Graph Reasoning (AGR), and Pathology-Aware Differential Refinement (PA-DFR) on top of a ConvNeXt backbone to improve representation learning for subtle and spatially heterogeneous radiographic findings.

    \item We introduce an evidential ordinal prediction design that combines Normal--Inverse-Gamma (NIG) regression with a pairwise ordinal ranking constraint, enabling continuous severity estimation together with an uncertainty surrogate for ambiguity-aware analysis.

    \item We provide empirical evaluation on both internal and external datasets, including comparisons with representative baselines, component-wise ablations, and qualitative analyses, to characterize the contribution of the proposed modules under the reported experimental protocol.
\end{itemize}

\begin{figure*}[!htbp]
  \centering
  \safeincludegraphics[width=\textwidth]{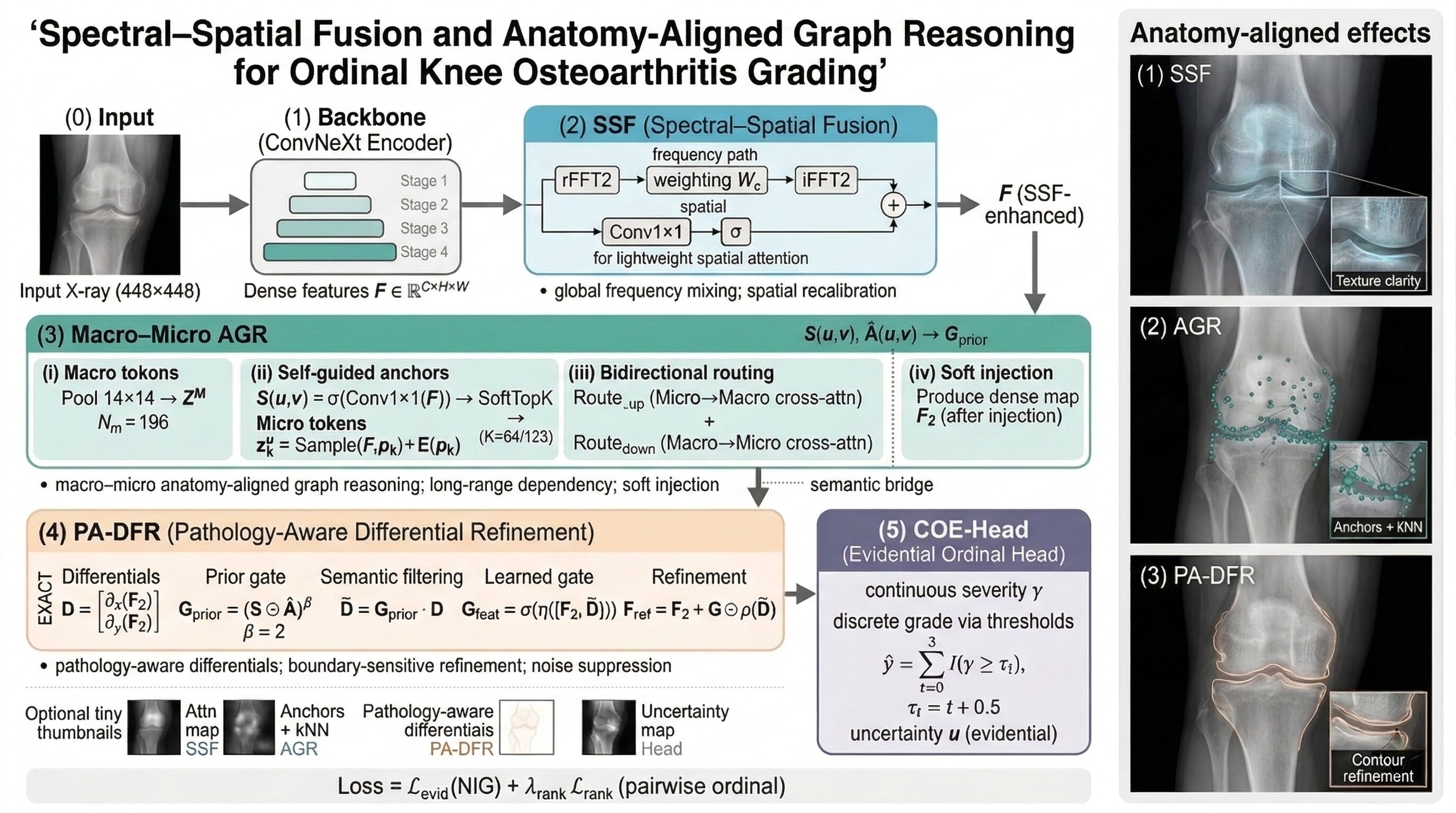}
  \caption{\textbf{Architectural overview of the proposed AGE-Net framework.} A ConvNeXt backbone initially extracts hierarchical semantic feature maps from the input radiograph. The backbone first extracts dense feature maps from the input radiograph, which are enhanced by SSF for frequency-sensitive micro-texture modulation. Subsequently, AGR is implemented as a macro--micro reasoning stage: a coarse $14\times14$ token lattice models global anatomical context, while a small set of learnable high-resolution anchors preserves micro-pathology evidence; bidirectional cross-attention routes information between anchor tokens and macro tokens prior to graph reasoning, and the refined anchor evidence is softly injected back into the dense feature map. Finally, PA-DFR performs semantically gated differential refinement using the internally predicted joint-interest map, selectively amplifying pathology-relevant boundary differentials while suppressing non-informative edges and noise. Global average pooling yields a condensed image-level embedding, which is projected into the COE-Head (Continuous Ordinal Evidential Head). The network is trained end-to-end, optimizing a synergistic loss function comprising evidential NIG regression and a strict pairwise ordinal ranking penalty.}
  \label{fig:overview}
\end{figure*}

\section{Related Work}

\subsection{Backbones for radiographs and medical imaging}
Convolutional neural networks (CNNs) have historically dominated musculoskeletal radiograph analysis, largely attributable to their robust inductive biases and optimization stability on datasets of constrained magnitude. While classic architectures—such as VGG\cite{simonyan2015vgg}, ResNet\cite{he2016resnet}, DenseNet\cite{huang2017densenet}, and Inception\cite{szegedy2016inception}—established foundational benchmarks, and EfficientNet subsequently refined accuracy-efficiency trade-offs via principled compound scaling\cite{zagoruyko2016wide}, ConvNeXt recently revolutionized the CNN landscape. By integrating micro-design principles derived from Vision Transformers (ViTs)\cite{vaswani2017attention}, ConvNeXt achieves remarkable transfer learning efficacy commensurate with transformer architectures, yet retains the spatial inductive biases crucial for radiographic interpretation.

Concurrently, Vision Transformers\cite{dosovitskiy2021vit} and their hierarchical derivatives\cite{keutayeva2024compact} \cite{roy2026osteotrans}, such as Swin Transformers\cite{liu2021swin}, excel in global context aggregation. However, their deployment in medical imaging is frequently compromised by an acute vulnerability to distribution shifts, hyperparameter sensitivity, and a voracious appetite for pretraining data—especially when labeled datasets are restricted. These considerations motivated our use of a modern CNN backbone (ConvNeXt) as the primary feature extractor for KL grading. Our goal was not to argue against transformer-based models in general, but rather to build upon a backbone with strong optimization stability and spatial inductive bias, and then augment it with modules designed to capture non-local anatomical reasoning and ordinal ambiguity in knee radiographs.

\subsection{Ordinal regression and ranking constraints}
KL grading is an intrinsically \emph{ordinal} endeavor \cite{chen2019fully} \cite{niu2016ordinal}: classification errors are fundamentally asymmetrical (e.g., misdiagnosing a Grade 4 instance as Grade 3 is clinically less catastrophic than classifying it as Gradeset 0), and adjacent grades are frequently demarcated by microscopic radiographic nuances. Consequently, formulating KL grading as a canonical, unconstrained multi-class classification problem engenders decision boundaries that blindly dismiss structural monotonicity, resulting in \emph{label-incoherent} predictions and algorithmic instability when confronting boundary cases characterized by high inter-reader variability.

A prevalent ordinal learning strategy mitigates this by fragmenting a $K$-grade hierarchy into $K\!-\!1$ ordered binary sub-tasks with shared latent representations, theoretically ensuring rank-consistent probabilities \cite{frank2001simple}. Architectures such as CORAL\cite{cao2020coral} operationalize this by optimizing a singular latent score across multiple ordered cutpoints, actively enforcing monotonic cumulative predictions. Conversely, direct regression to a continuous severity score followed by hard discretization, while computationally elegant, often underutilizes the discrete threshold taxonomy and exhibits vulnerability to heteroscedastic noise and extreme class imbalances\cite{gutierrez2015survey}.

As a highly synergistic alternative to threshold-based schema, \emph{pairwise ranking constraints} \cite{burges2005learning} explicitly enforce hierarchical integrity at the granular sample level. Given two discrete radiographs $(x_i, y_i)$ and $(x_j, y_j)$ where $y_i > y_j$, a ranking loss severely penalizes any violation of the ground-truth order by mandating that the predictive scalar $s(\cdot)$ satisfies $s(x_i) \ge s(x_j) + m$, with $m \ge 0$ functioning as the margin. Formulated as a hinge loss, $\mathcal{L}_{\mathrm{rank}}=\max\!\bigl(0,\, m - (s(x_i)-s(x_j))\bigr)$, As a complementary strategy to threshold-based ordinal modeling, pairwise ranking constraints enforce hierarchical consistency at the sample level. Given two radiographs $(x_i, y_i)$ and $(x_j, y_j)$ with $y_i > y_j$, a ranking loss encourages the predictive scalar $s(\cdot)$ to satisfy $s(x_i) \geq s(x_j) + m$, where $m \geq 0$ is a margin parameter. In hinge form, $\mathcal{L}_{\mathrm{rank}} = \max\!\left(0,\, m - (s(x_i)-s(x_j))\right)$. Such constraints provide a simple mechanism for encoding ordinal structure, selectively applying supervision to valid grade-separated pairs, and reducing boundary ambiguity through relative ordering. These properties make pairwise ranking a useful complement to evidential regression for KL grading.

\subsection{Graph reasoning for non-local dependencies}
Non-local structural dependencies dictate musculoskeletal radiographic interpretation: diagnostically critical cues (e.g., joint-space narrowing, marginal osteophytes, and subchondral sclerosis) are inherently spatially disseminated and mandate interpretation strictly relative to their overarching anatomical context, such as comprehensive tibiofemoral alignment. Graph neural networks (GNNs)\cite{bronstein2017geometric} \cite{kipf2017gcn} provide a mathematically elegant inductive bias for aggregating such distributed features, conceptualizing disparate regions or feature tokens as nodes and propagating synthesized context via dynamic message passing.

In advanced computer vision applications, dynamic graph topologies have proven exceptionally adept at mapping long-range, content-adaptive relationships. For instance, the EdgeConv mechanism within DGCNN dynamically recomputes graph neighborhoods within the latent feature space\cite{velickovic2018gat}, ensuring that connective topology organically evolves alongside the underlying representation to prioritize appearance-aware semantic interactions\cite{wang2019dgcnn}. Adapting these principles to graph-based reasoning over localized radiographic descriptors ensures that node interactions are governed by diagnostic similarity and structural proximity rather than static coordinate geometry\cite{qi2017pointnet}.

Within the specific purview of OA severity assessment, architectural graph reasoning is transformative. It explicitly facilitates: (i) the aggregation of subtle evidentiary cues across anatomically linked yet physically separated regions (e.g., the femoral condyle versus the tibial plateau); (ii) the definitive mapping of asymmetric compartmental pathologies (medial versus lateral degradation); and (iii) enhanced algorithmic resilience against positional and scale variance. These diagnostic imperatives directly inform the design philosophy of our Anatomical Graph Reasoner (AGR), which executes structured message passing over a condensed grid of radiographic tokens to codify non-local anatomical dialectics while retaining strict computational tractability.

\textbf{Relation to self-attention.}
In stark contrast to global self-attention mechanisms—which forcibly compute dense, all-to-all token interactions and exhibit severe regularization sensitivity in low-data medical paradigms—graph reasoning imposes an explicit, \emph{sparse} relational prior dictated by dynamically controlled local neighborhoods. This enforced sparsity dramatically curtails spurious long-range semantic mixing, markedly enhances structural interpretability at the discrete node/edge juncture, and delivers a computationally streamlined mechanism for isolating anatomically coherent non-local dependencies.

\subsection{Uncertainty estimation and evidential learning}
Rigorous uncertainty quantification is non-negotiable for clinical KL grading deployments, given the sheer prevalence of borderline cases where adjacent grades confound even specialized expert readers. Within this domain, it is analytically crucial to decouple \emph{aleatoric} uncertainty—originating from intrinsic label noise, subtle osteophytic ambiguity, or equivocal structural narrowing—from \emph{epistemic} uncertainty, which signifies fundamental model ignorance regarding out-of-distribution anatomical anomalies or rare pathological presentations\cite{kendall2017uncertainties}. High-fidelity uncertainty metrics directly empower dynamic triage protocols, selectively flagging low-confidence assessments for secondary human review and mitigating the devastating clinical consequences of overconfident algorithmic misdiagnosis.

While contemporary uncertainty paradigms—spanning Bayesian neural networks, Monte Carlo dropout formulations, deep model ensembles\cite{lakshminarayanan2017ensembles}\cite{gal2016dropout}, and post-hoc temperature calibration—provide robust theoretical foundations, their operational reliance on multiple stochastic forward passes or the maintenance of redundant model parameters precipitates unacceptable computational bottlenecks. Evidential deep learning presents a profoundly elegant resolution, synthesizing mathematically sound uncertainty estimates \emph{deterministically} within a singular network pass\cite{sensoy2018evidential}. Specifically, the Normal--Inverse-Gamma (NIG) evidential regression paradigm\cite{amini2020der} directly infers the hyper-parameters of an overarching predictive distribution alongside a scalar evidence quantity, robustly capturing complex, input-dependent uncertainty profiles without resorting to computationally ruinous sampling techniques\cite{malinin2018priornetworks}.

Furthermore, evidential formulations systematically couple predictive accuracy with uncertainty metrics via integrated evidence regularization: when a model generates an erroneous prediction or processes intrinsically ambiguous data, the loss function actively penalizes high evidence (thereby amplifying uncertainty), whereas confident predictions are strictly permitted only when unequivocally supported by the data manifold. These operational dynamics render evidential learning exceptionally optimized for KL grading, where clinical uncertainty is highly clustered precisely along the decision boundaries of adjacent severity levels.

\section{Method}
\label{sec:method}

\subsection{Task definition and outputs}
We formulate automated Kellgren--Lawrence (KL) grading from standard knee radiographs as the prediction of an ordinal target
$y \in \{0,1,2,3,4\}$.
Rather than treating the five grades as mutually independent classes, we represent disease severity with a continuous latent score
$\gamma \in \mathbb{R}$ that reflects the progressive nature of osteoarthritis.

\paragraph{Ordinal decision from a continuous score.}
The discrete prediction is obtained via an explicit ordinal-threshold rule,
\begin{equation}
\hat{y} \;=\; \sum_{t=0}^{3}\mathbb{I}\!\left(\gamma \ge \tau_t\right),
\qquad \tau_0 < \tau_1 < \tau_2 < \tau_3,
\label{eq:ordinal_threshold}
\end{equation}
where $\{\tau_t\}$ are cut-points separating adjacent KL levels. In our implementation, we use uniformly spaced thresholds $\tau_t = t + 0.5$, which is equivalent to rounding followed by clamping into $\{0,\dots,4\}$. Writing the decision in this form makes the ordinal mapping explicit and provides a convenient interface for discussing boundary proximity.

\paragraph{Uncertainty aligned with ordinal boundaries.}
In addition to $\gamma$, the model outputs an uncertainty estimate $u(\gamma)$ from the evidential head, intended to reflect predictive ambiguity. Because ordinal decisions are most sensitive near grade boundaries, we quantify boundary proximity by
\[
d(\gamma) = \min_{t\in\{0,1,2,3\}} |\gamma - \tau_t|.
\]
We interpret low-confidence cases as those exhibiting both small $d(\gamma)$ and large $u(\gamma)$, whereas high-confidence cases are expected to lie farther from decision boundaries with lower uncertainty. Under this formulation, the network is trained to regress accurate severity scores while exposing uncertainty that is informative for borderline cases. Consequently, the network is optimized to (i) regress accurate continuous severity scores $\gamma$ and (ii) produce uncertainty estimates that increase for anatomies near KL boundaries and decrease when radiographic evidence is definitive, enabling principled caution for borderline clinical evaluations.

\subsection{Overview of AGE-Net}
\label{sec:overview}
AGE-Net uses a convolutional backbone to extract a deep feature representation from the input radiograph, followed by three lightweight enhancement modules designed for KL grading: (i) Spectral--Spatial Fusion (SSF), which combines frequency-domain modulation with spatial recalibration; (ii) Anatomical Graph Reasoning (AGR), which models non-local anatomical dependencies through macro--micro relational reasoning; and (iii) Pathology-Aware Differential Refinement (PA-DFR), which emphasizes pathology-relevant boundary responses while suppressing less informative edges. The refined representation is then pooled into an image-level embedding and passed to an evidential regression head (COE-Head), which predicts a continuous severity score together with uncertainty-related parameters. During training, a pairwise ordinal ranking constraint is additionally applied to encourage monotonic consistency across severity levels.

\subsection{Backbone feature extraction}
\label{sec:backbone}
Given a preprocessed input radiograph $x \in \mathbb{R}^{3 \times H_0 \times W_0}$, the foundational ConvNeXt backbone\cite{liu2022convnext} generates a high-dimensional deep feature tensor:
\begin{equation}
F = f_{\mathrm{bb}}(x) \in \mathbb{R}^{C \times H \times W}.
\end{equation}
Crucially, the hierarchical spatial resolution of $F$ is strictly preserved prior to pooling, ensuring that highly localized and microscopically subtle osteophytic features are structurally retained for downstream enhancement.

\subsection{Spectral--Spatial Fusion (SSF)}
\label{sec:ssf}
\textbf{Rationale.}
Early-stage radiographic OA manifestations frequently present as nearly imperceptible, spatially constrained variations in trabecular texture and localized sclerosis. To optimally distinguish these high-frequency aberrations while concurrently retaining strict spatial selectivity, the SSF module harmonizes frequency-domain amplitude modulation with a tightly parameterized spatial gating mechanism\cite{chi2020fast}.

\textbf{Spectral modulation.}
Let $\mathcal{F}(\cdot)$ denote the 2D real Fast Fourier Transform (rFFT), utilized exclusively for real-valued spatial features. We first transition the feature map into the frequency domain:
\begin{equation}
\hat{F} = \mathcal{F}(F),
\end{equation}
subsequently applying a channel-wise, fully learnable complex scaling parameter $W_c \in \mathbb{C}$:
\begin{equation}
\hat{F}'_c = \hat{F}_c \odot W_c,\qquad
F_{\mathrm{spec}} = \mathcal{F}^{-1}(\hat{F}').
\end{equation}
This operation executes a global, channel-specific frequency recalibration, explicitly elevating the spectral coefficients synonymous with fine-grained, pathological radiographic micro-textures.

\textbf{Spatial recalibration.}
In parallel, a highly selective spatial gate $A = \sigma(g(F))$ is derived via a lightweight cascaded $1 \times 1$ convolutional bottleneck\cite{hu2018senet}:
\begin{equation}
A = \sigma\!\Bigl(\mathrm{Conv}_{1\times 1}\bigl(\phi(\mathrm{BN}(\mathrm{Conv}_{1\times 1}(F)))\bigr)\Bigr),
\end{equation}
where $\phi(\cdot)$ denotes a non-linear activation (e.g., ReLU) and $\sigma(\cdot)$ represents the sigmoid bounding function\cite{ioffe2015batchnorm}.

\textbf{Fusion.}
The final SSF output is synthesized via an additive residual stabilization strategy\cite{rao2021gfnet}, ensuring foundational feature preservation while mitigating vanishing gradients:
\begin{equation}
\mathrm{SSF}(F) = F + F_{\mathrm{spec}} + (F \odot A).
\end{equation}

\textbf{Computation.}
The SSF architecture introduces trivial parameter overhead (comprising solely a complex scaling vector per channel and dual $1\times 1$ point-wise convolutions). The rFFT operational cost scales gracefully as $O(CHW\log(HW))$ at the condensed post-backbone spatial resolution.

\begin{figure*}[!htbp]
  \centering
  \safeincludegraphics[width=\textwidth]{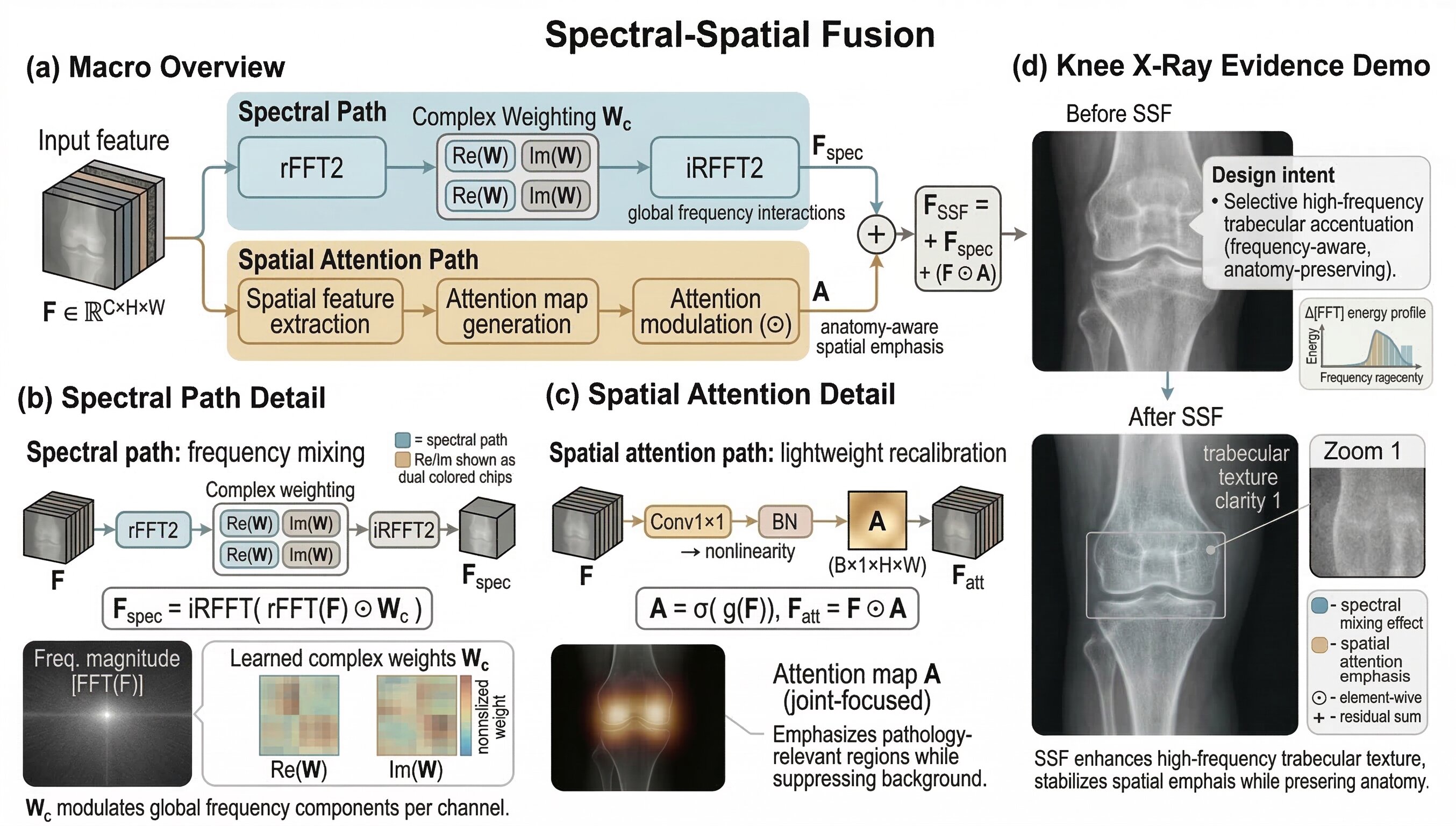}
  \caption{\textbf{Detailed schematic of the Spectral--Spatial Fusion (SSF) module.} The incoming deep feature map is bilaterally processed. The spectral pathway utilizes rFFT and iFFT transformations integrated with a learnable complex parameterization to actively amplify high-frequency osteophytic textures. Concurrently, the spatial pathway generates an adaptive recalibration gate. A final residual connection fuses these streams, preserving base semantics while heightening diagnostic sensitivity.}
  \label{fig:ssf}
\end{figure*}

\subsection{Anatomical Graph Reasoning (AGR): Macro--Micro Bidirectional Routing}

\textbf{Rationale.}
Early OA cues are subtle and spatially localized (e.g., marginal osteophytes and joint-space narrowing), while global anatomical context remains critical for robust KL grading.
Directly constructing graphs over dense pixels is computationally prohibitive and can dilute pathology with background.
We therefore adopt a \emph{macro--micro} design: a coarse token graph captures global context, while a small set of high-resolution \emph{anatomical anchors} preserves micro-pathology evidence. Crucially, anchors are routed \emph{into} the macro graph before reasoning, and refined macro context is routed back to anchors thereafter.

\textbf{Macro tokens.}
Given SSF-enhanced dense features $F\in\mathbb{R}^{C\times H\times W}$, we form macro tokens by adaptive pooling to a coarse lattice ($H_m\times W_m=14\times14$):
\begin{equation}
\begin{aligned}
F^{M} &= \mathrm{Pool}_{14\times14}(F)\in\mathbb{R}^{C\times H_m\times W_m}, \\
Z^{M} &= \{z_i^{M}\}_{i=1}^{N_m}, \quad N_m = H_mW_m .
\end{aligned}
\label{eq:macro_tokens}
\end{equation}

\textbf{Self-guided anchor sampling (micro tokens).}
To ensure stable convergence without hard-coded templates, we predict a soft joint-interest (saliency) map
\begin{equation}
S=\sigma(\mathrm{Conv}_{1\times1}(F))\in(0,1)^{H\times W},
\label{eq:saliency_map}
\end{equation}
and obtain $K$ anchor locations $P=\{p_k\}_{k=1}^{K}$ via a differentiable top-$K$ operator (e.g., SoftTopK / Gumbel-TopK) applied to $S$:
\begin{equation}
P=\mathrm{SoftTopK}(S;K),\qquad p_k=(u_k,v_k).
\label{eq:soft_topk}
\end{equation}
Anchor (micro) features are extracted at full resolution by differentiable sampling (bilinear grid sampling):
\begin{equation}
z_k^{\mu}=W_{\mu}\,\mathrm{Sample}(F,p_k)+E(p_k)\in\mathbb{R}^{d},\qquad
Z^{\mu}=\{z_k^{\mu}\}_{k=1}^{K}.
\label{eq:micro_tokens}
\end{equation}

\textbf{Bidirectional routing (Micro $\rightarrow$ Macro $\rightarrow$ Micro).}
Before graph reasoning, macro tokens absorb micro-pathology evidence via Micro$\rightarrow$Macro cross-attention:
\begin{equation}
\begin{aligned}
A^{\uparrow} &=
\operatorname{Softmax}\!\left(
\frac{Q(Z^{M})K(Z^{\mu})^{\top}}{\sqrt{d}} + B(\Pi,P)
\right), \\
\tilde z_i^{M} &=
z_i^{M} + \operatorname{MLP}\!\Big(
\big[z_i^{M},\ \sum\nolimits_{k}A^{\uparrow}_{ik}V(z_k^{\mu})\big]
\Big).
\end{aligned}
\label{eq:route_up}
\end{equation}
Here $A^{\uparrow}\in\mathbb{R}^{N_m\times K}$ and $\tilde Z^{M}=\{\tilde z_i^{M}\}_{i=1}^{N_m}$.

We then apply graph reasoning over macro tokens:
\begin{equation}
\bar Z^{M}=\mathrm{AGR}(\tilde Z^{M}).
\label{eq:agr_macro}
\end{equation}

Finally, refined macro context is routed back to anchors (Macro$\rightarrow$Micro):
\begin{equation}
\begin{aligned}
A^{\downarrow} &=
\operatorname{Softmax}\!\left(
\frac{Q(Z^{\mu})K(\bar Z^{M})^{\top}}{\sqrt{d}} + B(P,\Pi)
\right), \\
\bar z_k^{\mu} &=
z_k^{\mu} + \operatorname{MLP}\!\Big(
\big[z_k^{\mu},\ \sum\nolimits_{j}A^{\downarrow}_{kj}V(\bar z_j^{M})\big]
\Big).
\end{aligned}
\label{eq:route_down}
\end{equation}
Here $A^{\downarrow}\in\mathbb{R}^{K\times N_m}$ and $\bar Z^{\mu}=\{\bar z_k^{\mu}\}_{k=1}^{K}$.
Here $\Pi$ denotes macro-grid coordinates and $B(\cdot)$ is an optional relative-position bias.

\textbf{Soft injection back to dense space.}
The refined micro evidence $\bar Z^{\mu}$ is softly injected back to the dense feature map to form $F_2$ (details in Sec.~3.9), ensuring that high-frequency anchor cues influence subsequent boundary refinement and the final ordinal head.

\begin{figure*}[!htbp]
  \centering
  \safeincludegraphics[width=\textwidth]{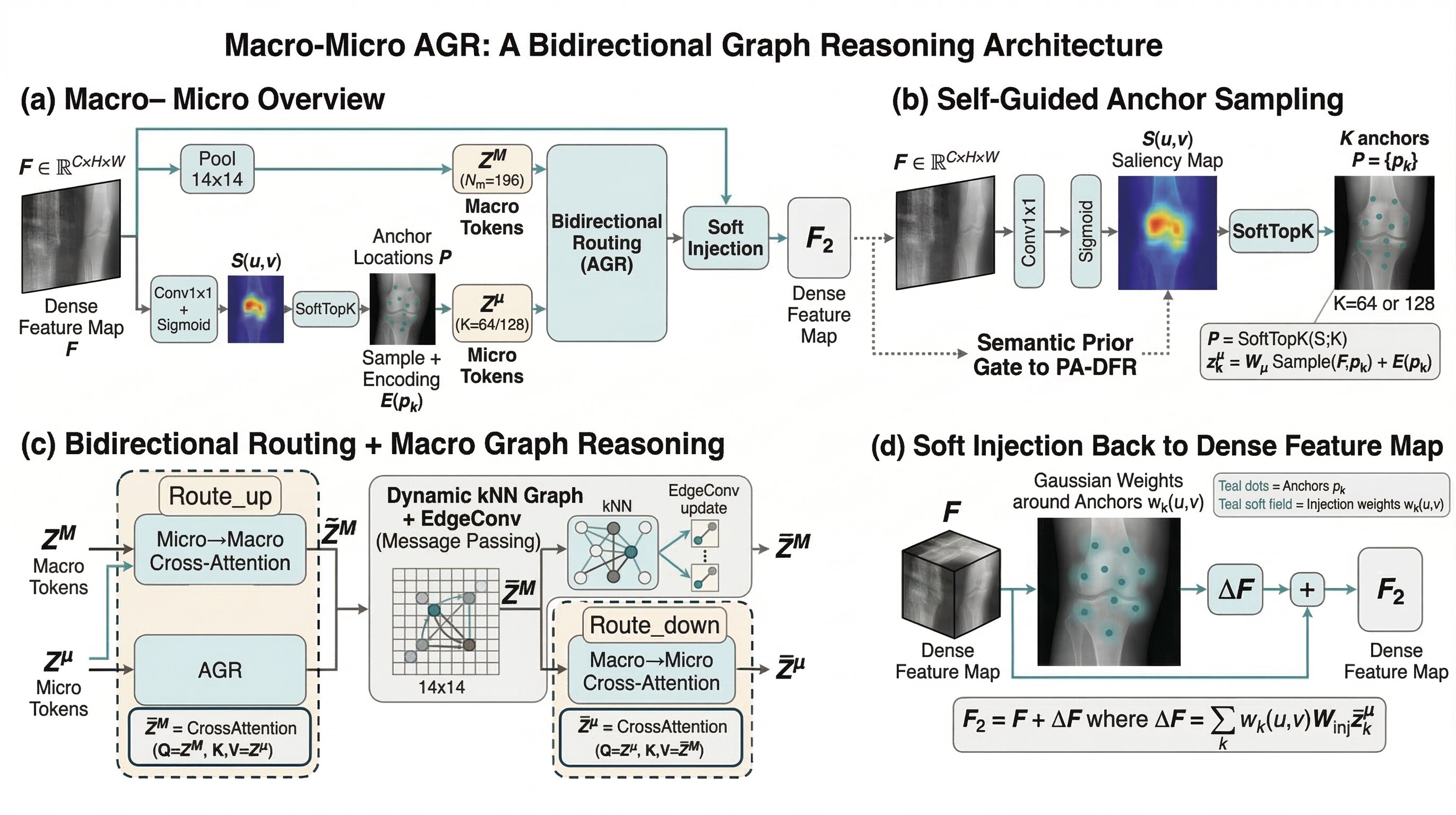}
  \caption{\textbf{Architecture of the Anatomical Graph Reasoner (AGR).} By projecting high-dimensional spatial maps into a compact, discretized token grid ($14 \times 14$), the AGR constructs a dynamic kNN graph that logically bridges anatomically relevant but spatially detached sub-regions (e.g., assessing symmetry between the medial and lateral tibiofemoral compartments). An EdgeConv-style message passing protocol aggregates these critical structural relationships, outputting a contextually enriched spatial attention gate that dynamically guides the primary feature stream via residual fusion.}
  \label{fig:agr}
\end{figure*}

\subsection{Pathology-Aware Differential Refinement (PA-DFR)}

\textbf{Rationale.}
Na\"ive differential operators indiscriminately respond to ubiquitous anatomical edges, soft-tissue boundaries, and noise, which can corrupt boundary-sensitive OA cues.
We therefore introduce a \emph{pathology-aware} differential refinement that leverages the internally predicted joint-interest map $S$ to gate differential responses, suppressing non-informative edges while amplifying pathology-relevant boundary cues.

\textbf{Differentials.}
Given the AGR-updated dense feature map $F_2\in\mathbb{R}^{C\times H\times W}$, we compute directional differentials (implemented as depthwise derivative filters):
\begin{equation}
D_x=\partial_x(F_2),\qquad D_y=\partial_y(F_2),\qquad D=[D_x,D_y].
\label{eq:differentials}
\end{equation}

\textbf{Semantic prior gating.}
We reuse the joint-interest map $S$ (Eq.~\eqref{eq:saliency_map}) as a spatial prior:
\begin{equation}
G_{\mathrm{prior}}(u,v)=S(u,v)^{\beta},\qquad \beta\ge 1,
\label{eq:prior_gate}
\end{equation}
and apply it to obtain semantically filtered differentials
\begin{equation}
\tilde D(u,v)=G_{\mathrm{prior}}(u,v)\cdot D(u,v).
\label{eq:semantic_diff}
\end{equation}

\textbf{Feature-adaptive gating and residual refinement.}
A learnable gate further modulates the filtered differentials:
\begin{equation}
\begin{aligned}
G_{\mathrm{feat}} &= \sigma\!\big(\eta([F_2,\tilde D])\big)\in(0,1)^{C\times H\times W},\\
G &= \mathrm{clip}\big(G_{\mathrm{prior}}\cdot \mathbf{1}_C \odot G_{\mathrm{feat}},\,0,\,1\big).
\end{aligned}
\label{eq:full_gate}
\end{equation}
where $\eta(\cdot)$ is a lightweight $1\times1$ bottleneck and $\mathbf{1}_C$ denotes broadcasting $G_{\mathrm{prior}}$ across channels.
The final refinement is a gated residual update:
\begin{equation}
F_{\mathrm{ref}}=F_2 + G\odot \rho(\tilde D),
\label{eq:pa_dfr}
\end{equation}
with $\rho(\cdot)$ a linear projection for channel alignment.
This construction makes PA-DFR near-identity outside the joint ROI (where $S\approx0$), preventing indiscriminate edge amplification while sharpening pathology-relevant boundary transitions.

\begin{figure*}[!htbp]
  \centering
  \safeincludegraphics[width=\textwidth]{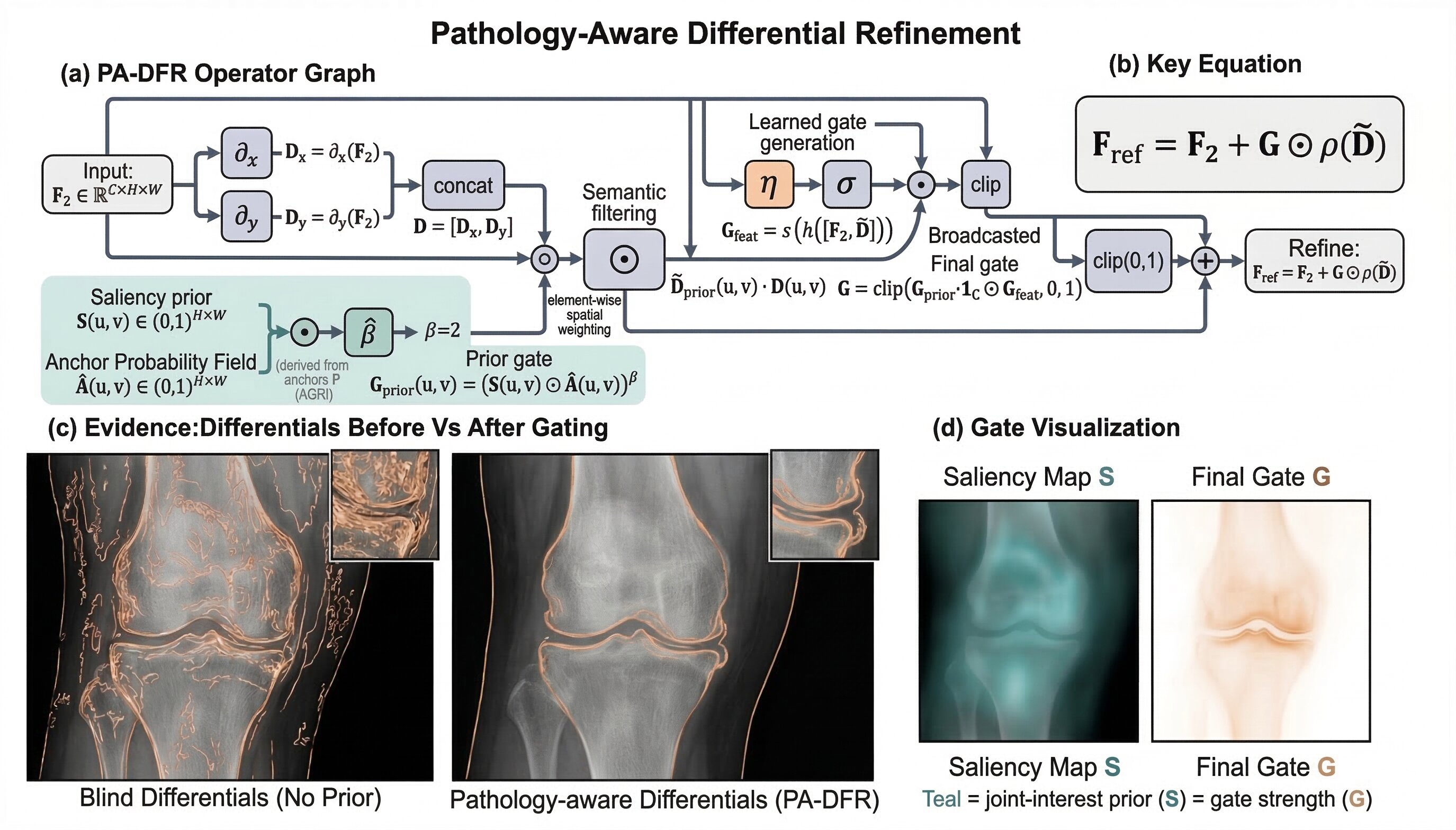}
  \caption{\textbf{Operational flow of the Differential Refiner (DFR) module.} To explicitly delineate defining structural degenerations—specifically asymmetrical joint-space narrowing and severe marginal bone attrition—DFR calculates local semantic gradients utilizing a $3 \times 3$ depthwise convolutional kernel. The resulting absolute differential response map sharply emphasizes osteochondral borders, which are subsequently integrated back into the primary diagnostic representation via a SiLU activation within a stabilizing residual pathway.}
  \label{fig:dfr}
\end{figure*}

\subsection{Global pooling and prediction}
The global pooling operator compresses the spatial tensor into a dense image-level embedding. This embedding is ingested by an evidential regression head (\textbf{COE-Head}), which concurrently predicts the continuous severity score and its associated uncertainty parameters. During the optimization phase, a supplementary pairwise ordinal ranking constraint is applied to rigorously enforce monotonicity across the predicted severity continuum.

\begin{figure*}[!htbp]
  \centering
  \safeincludegraphics[width=\textwidth]{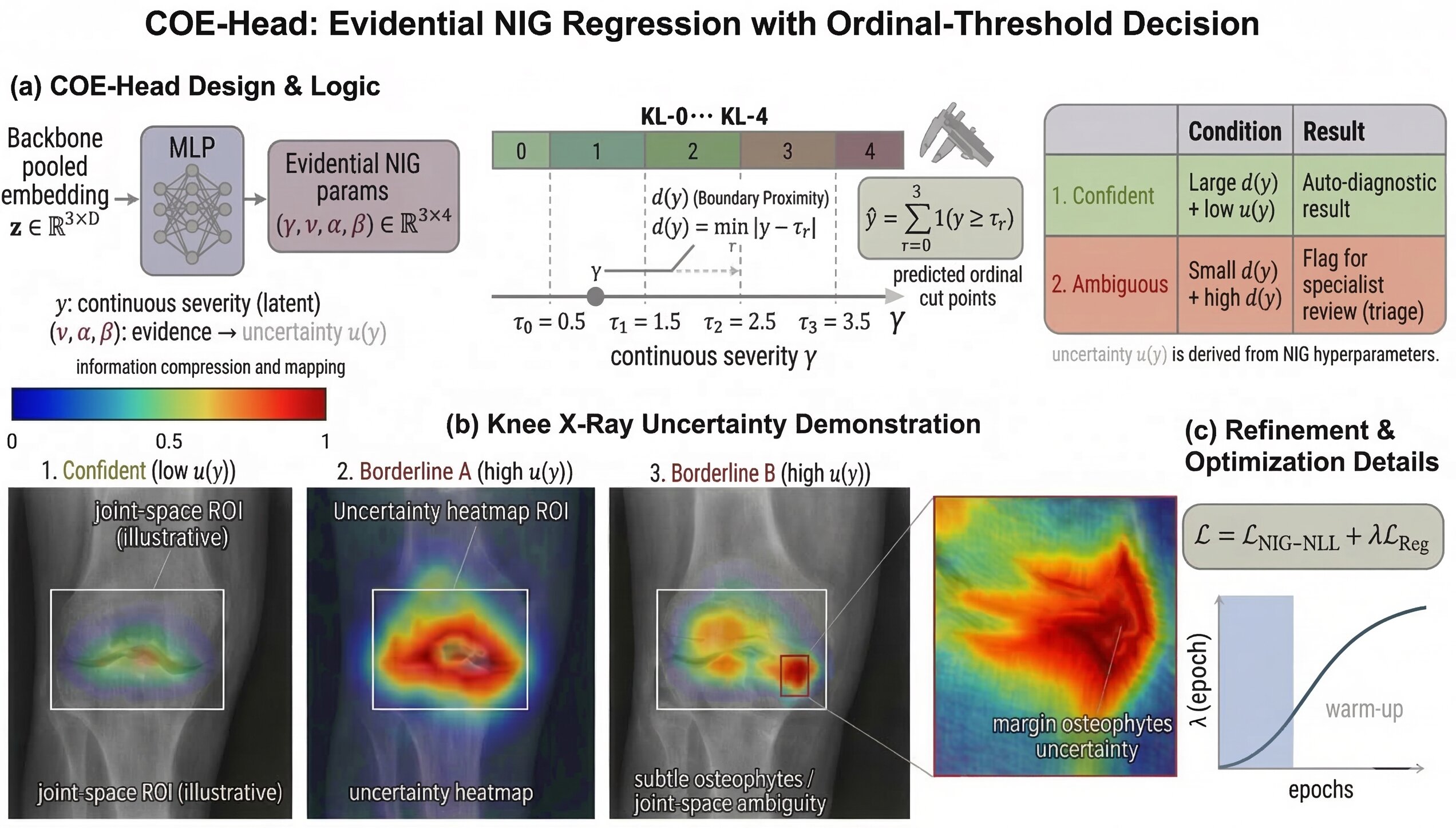}
  \caption{Architecture of the COE-Head (Evidential NIG Regression). The globally pooled embedding is mapped to the hyper-parameters $(\gamma,\nu,\alpha,\beta)$ of a Normal--Inverse-Gamma distribution. This deterministic formulation simultaneously estimates a continuous severity score and an uncertainty-related quantity from the predicted evidence parameters, without requiring stochastic sampling.}
  \label{fig:coe}
\end{figure*}

\subsection{Ordinal ranking constraint}
\label{sec:rank}
\paragraph{Rationale.}
To better preserve ordinal consistency in KL grading, we introduce a pairwise ranking constraint during training. This auxiliary objective penalizes violations of the expected severity ordering and complements the evidential regression objective by emphasizing relative ordering between grade-separated samples.

\textbf{Valid pairs and margin.}
During mini-batch processing, we isolate all structurally valid, grade-separated image pairs:
\begin{equation}
\Omega=\{(i,j)\mid y_i-y_j\ge 1\}.
\end{equation}
Given a pre-established severity margin $m=0.8$, the ordinal ranking penalty is explicitly defined as:
\begin{equation}
\mathcal{L}_{\mathrm{rank}}=\frac{1}{|\Omega|}\sum_{(i,j)\in\Omega}\max(0, m-(\gamma_i-\gamma_j)).
\end{equation}
Crucially, this ranking loss is conditionally deactivated whenever Mixup augmentation is triggered, as the interpolation of labels intrinsically invalidates the discrete pairwise inequality constraint.

\textbf{Total objective.}
The complete end-to-end optimization objective is formulated as:
\begin{equation}
\mathcal{L}=\mathcal{L}_{\mathrm{evi}}+\alpha_{\mathrm{rank}}\mathcal{L}_{\mathrm{rank}}
+\lambda_{\mathrm{div}}\mathcal{L}_{\mathrm{div}}+\lambda_{\mathrm{att}}\mathcal{L}_{\mathrm{att}},
\label{eq:total_loss}
\end{equation}
where $\mathcal{L}_{\mathrm{evi}}$ and $\mathcal{L}_{\mathrm{rank}}$ follow Eq.~(7), and the latter is conditionally deactivated when Mixup is applied (as interpolated labels invalidate discrete pairwise inequalities).
To stabilize self-guided anchor sampling, we employ (i) a diversity regularizer preventing anchor collapse,
$\mathcal{L}_{\mathrm{div}}=\frac{1}{K(K-1)}\sum_{k\neq \ell}\exp\!\left(-\|p_k-p_\ell\|_2^2/(2\sigma_p^2)\right)$,
and (ii) an attention-alignment term encouraging anchors to reside in high-interest regions,
$\mathcal{L}_{\mathrm{att}}=-\frac{1}{K}\sum_{k}\log S(p_k)$.

\begin{figure*}[!htbp]
  \centering
  \safeincludegraphics[width=\textwidth]{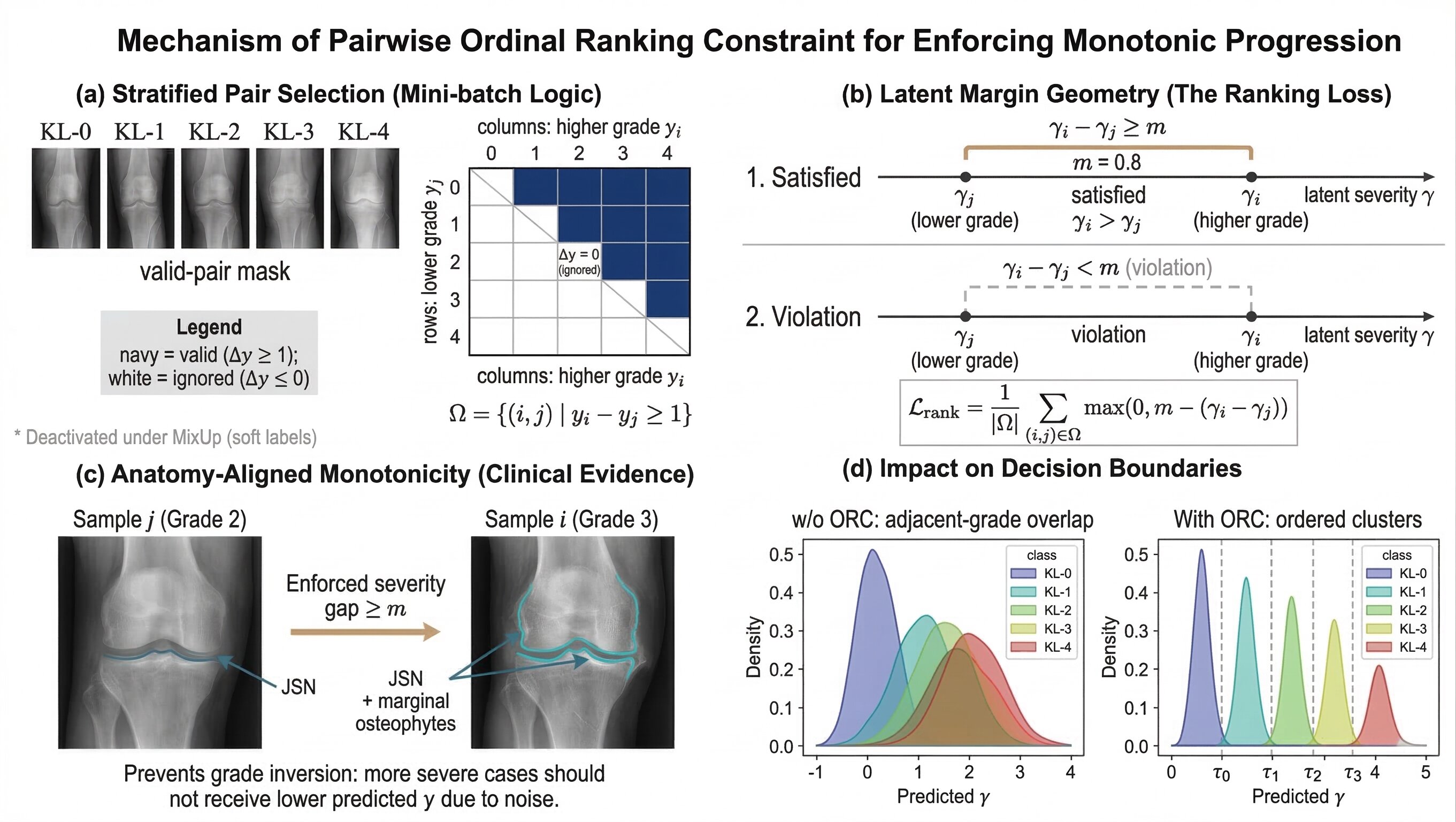}
  \caption{\textbf{Operational logic of the Ordinal Ranking Constraint.} The auxiliary ranking loss systematically evaluates valid, grade-separated sample pairs within each mini-batch. By strictly enforcing a predefined margin ($m=0.8$) between the predicted continuous severity scores of disparate grades, the framework irrevocably binds the optimization landscape to the monotonic, progressive reality of osteoarthritis degeneration.}
  \label{fig:rank}
\end{figure*}

\subsection{End-to-end inference pipeline}
\label{sec:pipeline}
The sequential inference cascade initiates with foundational feature extraction
$F_0=f_{\mathrm{bb}}(x)$, and proceeds as:
\begin{align}
F_1 &= \mathrm{SSF}(F_0), \nonumber\\
(Z^{M},Z^{\mu},P,S) &= \mathrm{AnchorSampler}(F_1), \nonumber\\
\tilde Z^{M} &= \mathrm{Route}_{\uparrow}(Z^{M},Z^{\mu},\Pi,P), \nonumber\\
\bar Z^{M} &= \mathrm{AGR}(\tilde Z^{M}), \nonumber\\
\bar Z^{\mu} &= \mathrm{Route}_{\downarrow}(Z^{\mu},\bar Z^{M},P,\Pi), \nonumber\\
F_2 &= \mathrm{Inject}(F_1,\bar Z^{\mu},P), \nonumber\\
F_3 &= \mathrm{PA\text{-}DFR}(F_2;S).
\label{eq:inference_pipeline}
\end{align}
The fully enhanced tensor $F_3$ undergoes global pooling to synthesize $z$, which structurally drives the COE-Head prediction of $(\gamma,\nu,\alpha,\beta)$. During clinical inference, the resultant continuous predictive scalar $\gamma$ is categorically mapped to a definitive discrete KL grade strictly via mathematical rounding and rigid clamping to the valid domain $\{0,\ldots,4\}$. During our comprehensive ablation iterations, targeted modules are systematically excised from this sequential cascade while holding all complementary components and optimization hyperparameters utterly constant.

\section{Experiments}

\subsection{Dataset and evaluation protocol}
\begin{figure*}[!htbp]
\centering
\safeincludegraphics[width=\textwidth]{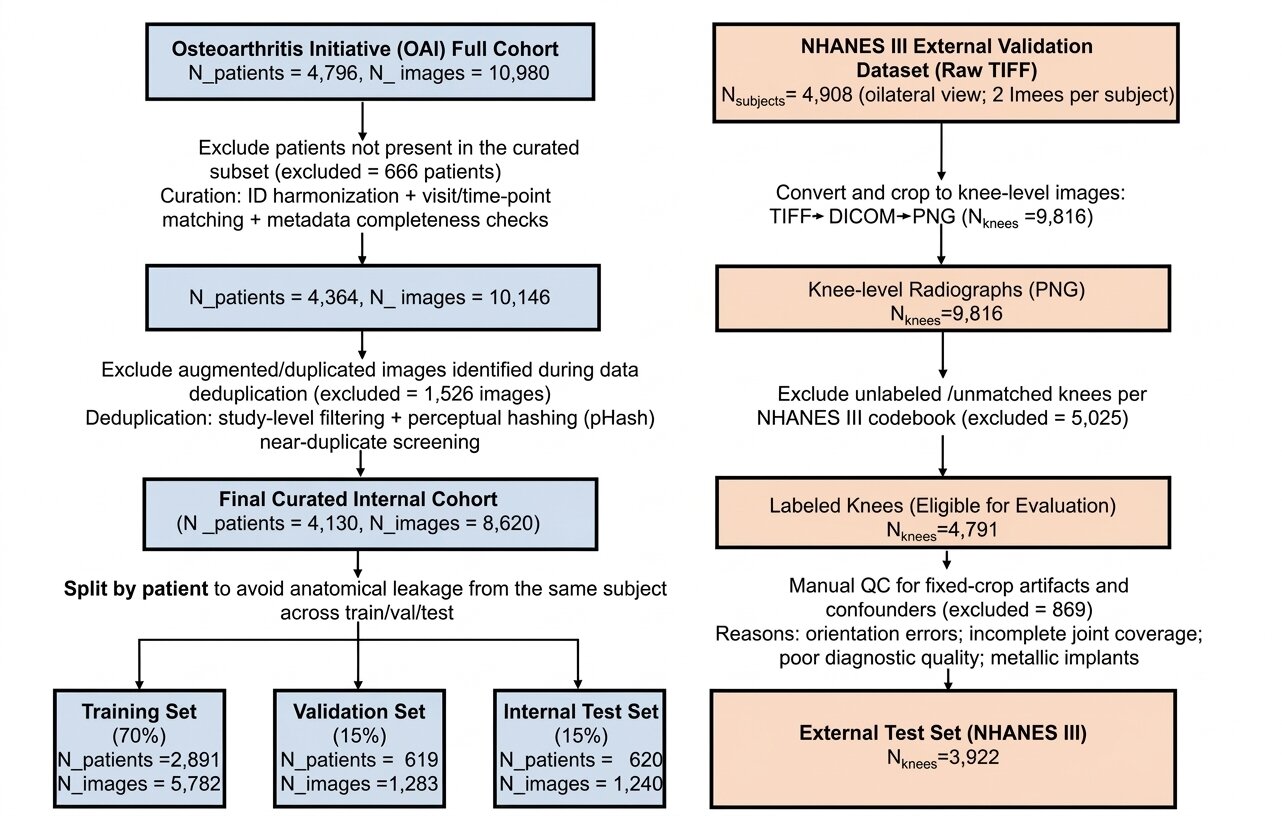}
\caption{\textbf{Exhaustive dataset curation and segregation protocol.} (Left) The algorithmic flow delineating the filtering, exclusion, and strict patient-level train/validation/test segregation parameters executed upon the massive internal OAI cohort. (Right) The parallel preprocessing, standardization, and manual radiological Quality Control (QC) pipeline utilized to forge the completely independent NHANES III external validation cohort.}
\label{fig:datafiltering}
\end{figure*}

We constructed an internal development cohort from the Osteoarthritis Initiative (OAI)\cite{peterfy2008oai} and used NHANES III as an external evaluation cohort (Fig.~\ref{fig:datafiltering}). The OAI cohort initially included 4,796 patient records comprising 10,980 knee radiographs. After filtering patients outside the target diagnostic subset and removing duplicated or augmented artifacts, the final internal cohort contained 4,130 unique patients and 8,620 radiographs. To avoid subject-level leakage, the internal data were partitioned at the patient level into training, validation, and testing subsets (2,891, 619, and 620 patients, corresponding to 5,782, 1,238, and 1,240 radiographs, respectively).

For NHANES III, raw TIFF images were processed into knee-level PNG images, yielding 9,816 isolated knees. After excluding unlabeled or diagnostically unmatched entries (5,025 exclusions) and removing severely degraded or improperly aligned images through manual quality control (869 exclusions), the final external test set comprised 3,922 knees. All labels followed the five-grade KL scale $\{0,1,2,3,4\}$. Unless otherwise stated, all experiments were repeated across three random initialization seeds and are reported as mean $\pm$ standard deviation.

For the external NHANES III validation cohort, raw archival TIFF files were meticulously processed and cropped into distinct knee-level PNG formats (totaling 9,816 isolated knees). Following the stringent exclusion of unlabeled or diagnostically unmatched entries (5,025 exclusions) and the manual radiological purging of highly degraded or improperly aligned images via strict Quality Control (QC) standards (869 exclusions), we finalized a remarkably robust external test set of 3,922 discrete knees. All targeted KL diagnostic labels adhere strictly to the standardized five-tier ordinal scale $\{0,1,2,3,4\}$. To ensure absolute statistical rigor, all experimental runs are executed across three totally divergent random initialization seeds, reporting explicit mean metrics accompanied by comprehensive standard deviation bounds. Detailed demographic and baseline clinical characteristics characterizing the patient populations are comprehensively enumerated in Table \ref{tab:patient_stats}.

\ModuleFloatBarrier

\begin{table*}[!htbp]
\centering
\caption{\textbf{Comprehensive patient demographics and baseline clinical characteristics.}}
\label{tab:patient_stats}
\rmfamily\small
\renewcommand{\arraystretch}{0.95}
\begin{tabular*}{\textwidth}{@{\extracolsep{\fill}}lccccc}
\toprule
Characteristic & Overall & Training & Validation & Testing & P-value* \\
\midrule
Age, years (Mean $\pm$ SD) & 60.91 $\pm$ 9.14 & 61.04 $\pm$ 9.18 & 60.64 $\pm$ 9.13 & 60.54 $\pm$ 8.94 & 0.2583 \\
Sex (Female), n (\%) & 2397 (58.0\%) & 1681 (58.1\%) & 341 (55.1\%) & 375 (60.5\%) & 0.3042 \\
BMI, kg/m$^2$ (Mean $\pm$ SD) & 28.51 $\pm$ 4.78 & 28.50 $\pm$ 4.75 & 28.53 $\pm$ 4.85 & 28.55 $\pm$ 4.87 & 0.8497 \\
\midrule
Total images, n & 8260 & 5782 & 1238 & 1240 & -- \\
KL Grade 0, n (\%) & 3253 (39.4\%) & 2228 (38.5\%) & 510 (41.2\%) & 515 (41.5\%) & 0.0534 \\
KL Grade 1, n (\%) & 1495 (18.1\%) & 1056 (18.3\%) & 218 (17.6\%) & 221 (17.8\%) & 0.7454 \\
KL Grade 2, n (\%) & 2175 (26.3\%) & 1568 (27.1\%) & 293 (23.7\%) & 314 (25.3\%) & 0.2075 \\
KL Grade 3, n (\%) & 1086 (13.1\%) & 761 (13.2\%) & 174 (14.1\%) & 151 (12.2\%) & 0.3741 \\
KL Grade 4, n (\%) & 251 (3.0\%) & 169 (2.9\%) & 43 (3.5\%) & 39 (3.1\%) & 0.7439 \\
\bottomrule
\multicolumn{6}{l}{\rmfamily\small
\renewcommand{\arraystretch}{0.95} *P-value mathematically quantifies the distributional discrepancy between the Training and Testing subsets.}
\end{tabular*}
\end{table*}
\ModuleFloatBarrier

\subsection{Implementation details}
Input radiographs were first resized to $512 \times 512$ pixels and then randomly cropped to $448 \times 448$ during training. The model was optimized end-to-end for 300 epochs using AdamW\cite{loshchilov2019decoupled} with a cosine annealing learning rate schedule\cite{loshchilov2017sgdr}. Data augmentation included RandomAffine, ColorJitter, Mixup\cite{zhang2018mixup} (applied with probability $p=0.5$), and Random Erasing\cite{zhong2020random} (applied with probability $p=0.25$). Mixed-precision training was used for efficiency. We also maintained an exponential moving average (EMA) of model weights with decay 0.999 during training. At inference time, test-time augmentation (TTA) averaged predictions from the original image and its horizontal flip. Additional implementation details are summarized in Table\ref{tab:impl}.

\begin{table}[!htbp]
\centering
\caption{\textbf{Exhaustive overview of network training and optimization configurations.}}
\label{tab:impl}
\rmfamily\small
\renewcommand{\arraystretch}{0.95}
\setlength{\tabcolsep}{5pt}
\begin{tabular}{p{0.40\linewidth} p{0.54\linewidth}}
\toprule
Optimization Parameter & Assigned Configuration \\
\midrule
Primary Backbone Base & ConvNeXt-Base \cite{liu2022convnext} (Pre-initialized via offline generic vision weights) \\
Ingested Spatial Resolution & $448\times448$ pixels \\
Total Optimization Epochs & 300 cycles \\
Weight Update Algorithm & AdamW \cite{loshchilov2019decoupled} \\
Baseline Learning Rate (LR) & $1\times10^{-5}$ (For customized module heads), Backbone scaled tightly to $0.2\times$ \\
LR Decay Trajectory & Cosine annealing protocol \cite{loshchilov2017sgdr} \\
Global Weight Decay & $1\times10^{-5}$ \\
Weight Smoothing (EMA) & Exponential decay rate of 0.999 \\
Mixup Augmentation & Applied continuously with $p=0.5$ \cite{zhang2018mixup} \\
Random Erasing Augmentation & Applied continuously with $p=0.25$ \cite{zhong2020random} \\
Test-Time Augmentation (TTA) & Symmetrical horizontal flip averaging \\
Ordinal Ranking Hyperparameters & Penalty scalar $\alpha_{\text{rank}}=2.0$, strict enforced margin $m=0.8$ \\
\bottomrule
\end{tabular}
\end{table}
\ModuleFloatBarrier

\begin{table*}[!htbp]
\centering
\caption{\textbf{Comprehensive benchmark evaluation against state-of-the-art diagnostic paradigms (Results aggregated across three independent initialization seeds, expressed as Mean $\pm$ Standard Deviation).}}
\label{tab:main}
\rmfamily\small
\renewcommand{\arraystretch}{0.92}
\setlength{\tabcolsep}{3.5pt}
\begin{tabular}{lccccc}
\toprule
Target Architecture & QWK$\uparrow$ & MSE$\downarrow$ & ACC$\uparrow$ & Sensitivity$\uparrow$ & Specificity$\uparrow$ \\
\midrule
\textbf{AGE-Net (Proposed Full Framework)} & \textbf{0.9017$\pm$0.0045} & \textbf{0.2349$\pm$0.0028} & \textbf{0.7839$\pm$0.0056} & \textbf{0.8500$\pm$0.0096} & \textbf{0.8676$\pm$0.0115} \\
\midrule
VGG16 \cite{simonyan2015vgg} & 0.8654$\pm$0.0069 & 0.3046$\pm$0.0102 & 0.7154$\pm$0.0089 & 0.8012$\pm$0.0198 & 0.8705$\pm$0.0166 \\
EfficientNet \cite{tan2019efficientnet} & 0.8638$\pm$0.0012 & 0.3078$\pm$0.0060 & 0.7087$\pm$0.0049 & 0.8444$\pm$0.0222 & 0.8471$\pm$0.0244 \\
ConvNeXt-Base \cite{liu2022convnext} (Raw Baseline) & 0.8602$\pm$0.0077 & 0.3248$\pm$0.0118 & 0.7132$\pm$0.0087 & 0.7944$\pm$0.0259 & 0.9029$\pm$0.0311 \\
Inception-V3 \cite{szegedy2016inception} & 0.8475$\pm$0.0075 & 0.3252$\pm$0.0069 & 0.6752$\pm$0.0030 & 0.8401$\pm$0.0186 & 0.8286$\pm$0.0239 \\
DenseNet121 \cite{huang2017densenet} & 0.8427$\pm$0.0088 & 0.3274$\pm$0.0032 & 0.6629$\pm$0.0126 & 0.8358$\pm$0.0166 & 0.8419$\pm$0.0167 \\
ResNet50 \cite{he2016resnet} & 0.8271$\pm$0.0012 & 0.3711$\pm$0.0021 & 0.6272$\pm$0.0033 & 0.8324$\pm$0.0075 & 0.8357$\pm$0.0100 \\
\midrule
Swin-Transformer (Tiny) \cite{liu2021swin} & 0.8537$\pm$0.0014 & 0.3562$\pm$0.0083 & 0.6777$\pm$0.0025 & 0.8106$\pm$0.0746 & 0.8442$\pm$0.0781 \\
Vision Transformer (Base) \cite{dosovitskiy2021vit} & 0.8431$\pm$0.0014 & 0.3692$\pm$0.0060 & 0.6575$\pm$0.0047 & 0.7748$\pm$0.0229 & 0.8846$\pm$0.0176 \\
Compact Convolutional Transformer-14 \cite{keutayeva2024compact} & 0.8407$\pm$0.0029 & 0.3700$\pm$0.0072 & 0.6513$\pm$0.0049 & 0.8055$\pm$0.0201 & 0.8535$\pm$0.0201 \\
DeiT (Small) \cite{jahan2024koa} & 0.8342$\pm$0.0026 & 0.3825$\pm$0.0125 & 0.6462$\pm$0.0026 & 0.7605$\pm$0.0201 & 0.8766$\pm$0.0210 \\
\bottomrule
\end{tabular}
\end{table*}
\ModuleFloatBarrier

\subsection{Metrics}
Model performance was evaluated using Quadratic Weighted Kappa (QWK)\cite{cohen1968weighted}, Mean Squared Error (MSE), Top-1 Accuracy (ACC), Macro-F1 score, and Recall. For categorical metrics such as QWK, ACC, F1, and Recall, the continuous severity predictions were rounded to the nearest integer and clamped to the valid KL range $[0,4]$.

\section{Results}

\subsection{Main comparison}
The benchmark results in Table\ref{tab:main} show that AGE-Net achieved the strongest overall performance among the evaluated models under the reported experimental protocol. Specifically, AGE-Net obtained a QWK of $0.9017 \pm 0.0045$ and an MSE of $0.2349 \pm 0.0028$ on the internal test set. It also outperformed the raw ConvNeXt-Base backbone and the other representative baselines included in this study. These results suggest that the proposed combination of anatomy-aware refinement and ordinal evidential prediction is effective for KL grading on the evaluated dataset.

\begin{figure*}[!htbp]
\centering
\safeincludegraphics[width=0.95\textwidth]{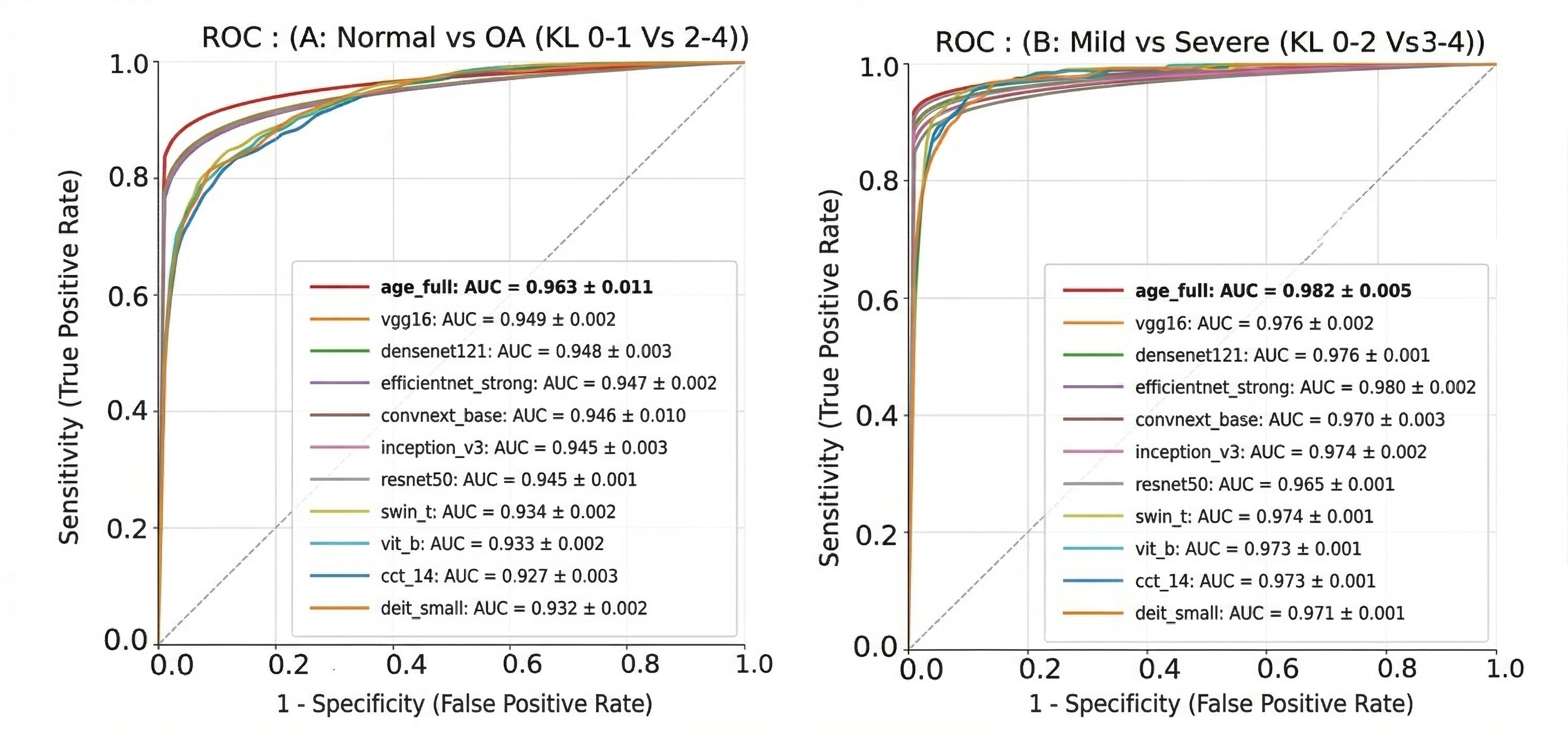}
\caption{\textbf{Receiver Operating Characteristic (ROC) analytics quantifying diagnostic efficacy.} Curve A maps the binary capability of the network to reliably screen for generalized Osteoarthritis presence (achieving a formidable AUC of 0.936). Curve B delineates the significantly more complex task of multi-class progressive severity grading (peaking at an AUC of 0.982). All plotted trajectories represent mathematically averaged distributions spanning 3 independent optimization cycles; shaded localized regions visually denote precise standard deviation intervals.}
\label{fig:roc}
\end{figure*}

\subsection{Ablation study}
Table\ref{tab:abl} summarizes the ablation results for the main components of AGE-Net. Removing the ordinal ranking constraint led to a reduction in QWK and an increase in MSE, indicating that explicit ordinal supervision contributes to prediction quality. Likewise, removing AGR degraded performance, suggesting that modeling non-local anatomical interactions is beneficial for KL grading. Larger drops were observed when SSF, PA-DFR, or the COE-Head were removed, supporting the complementary roles of frequency-aware enhancement, boundary-aware refinement, and evidential ordinal prediction in the overall framework.

\begin{table*}[!htbp]
\centering
\caption{\textbf{Architectural ablation study quantifying the isolated impact of integrated modules (Results derived across 3 independent seeds).}}
\label{tab:abl}
\rmfamily\normalsize
\begin{tabular*}{\textwidth}{@{\extracolsep{\fill}}lcc}
\toprule
Model Variant Architecture & Computed QWK$\uparrow$ & Resultant MSE$\downarrow$ \\
\midrule
\textbf{AGE-Net (Fully Intact Framework)} & \textbf{0.9017$\pm$0.0045} & \textbf{0.2349$\pm$0.0028} \\
w/o Ordinal Ranking Constraint & 0.8963$\pm$0.0033 & 0.2600$\pm$0.0083 \\
w/o Anatomical Graph Reasoner (AGR) & 0.8932$\pm$0.0018 & 0.2665$\pm$0.0017 \\
Bare ConvNeXt Baseline Structure & 0.8602$\pm$0.0077 & 0.3248$\pm$0.0118 \\
\midrule
\textit{w/o Spectral--Spatial Fusion (SSF)} & 0.8229$\pm$0.0132 & 0.3950$\pm$0.0136 \\
\textit{w/o Differential Refinement (DFR)} & 0.8289$\pm$0.0014 & 0.3908$\pm$0.0113 \\
\textit{w/o Continuous Ordinal Evidential (COE) Head} & 0.8375$\pm$0.0032 & 0.3586$\pm$0.0053 \\
\bottomrule
\end{tabular*}
\end{table*}

\subsection{Per-grade analysis and interpretability}
\begin{figure*}[!htbp]
\centering
\safeincludegraphics[width=\textwidth]{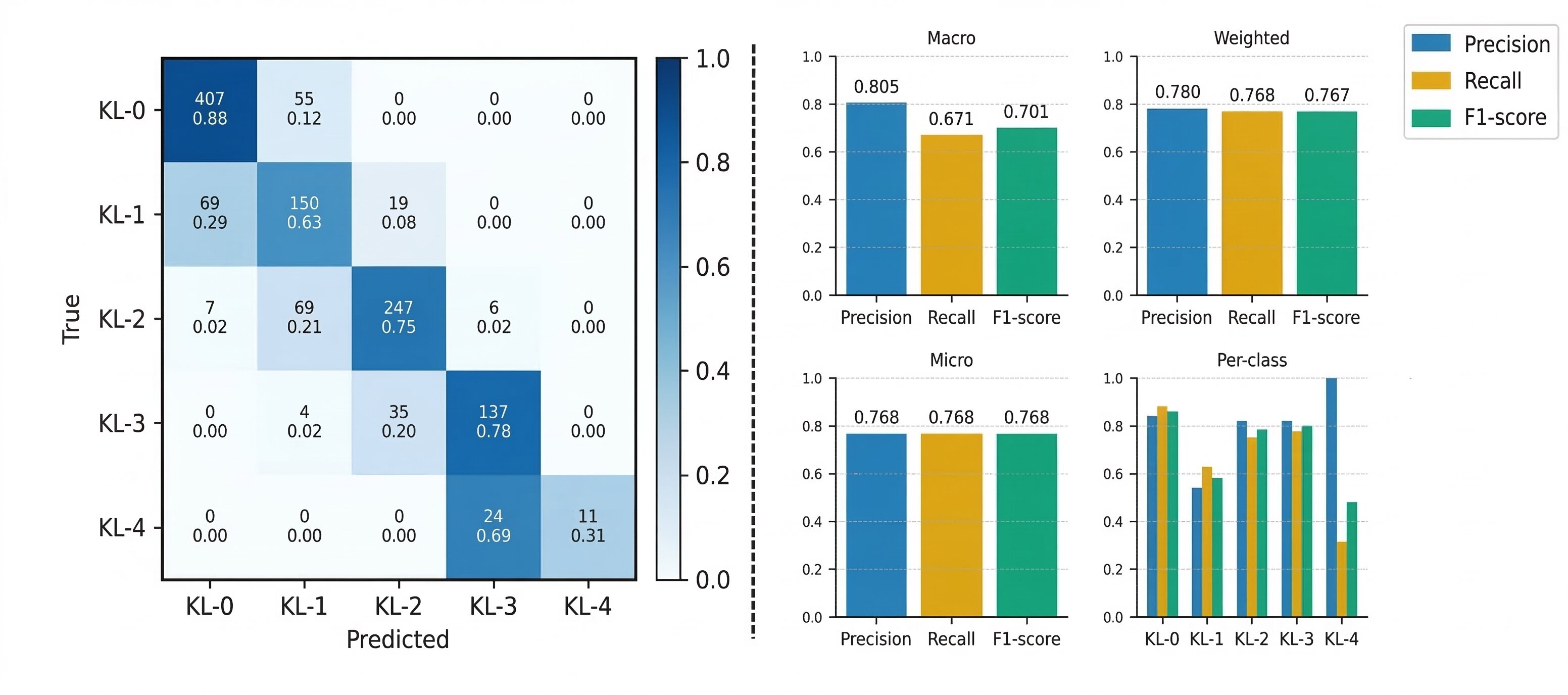}
\caption{\textbf{Row-normalized confusion matrix executing precise KL grading via AGE-Net.} The matrix clearly delineates that the overwhelming majority of classification divergence occurs strictly between adjacent, structurally similar grades (e.g., differentiating Grade 2 from Grade 3), directly reflecting the natural clinical ambiguity of progressive morphological transitions rather than systemic structural network failures.}
\label{fig:confusion}
\end{figure*}

Figure\ref{fig:confusion} shows the row-normalized confusion matrix for AGE-Net. Most misclassifications were concentrated between adjacent grades, which is consistent with the gradual nature of KL progression and the known ambiguity of borderline cases. Figures~10 and~11 provide qualitative visualization of model attention using gradient-based attribution. These examples suggest that the model tends to focus on clinically relevant joint regions, such as areas associated with joint-space narrowing and osteophyte formation, when producing KL predictions. However, these visualizations should be interpreted as qualitative evidence rather than definitive proof of mechanistic interpretability.

\begin{figure*}[!htbp]
\centering
\safeincludegraphics[width=0.9\textwidth]{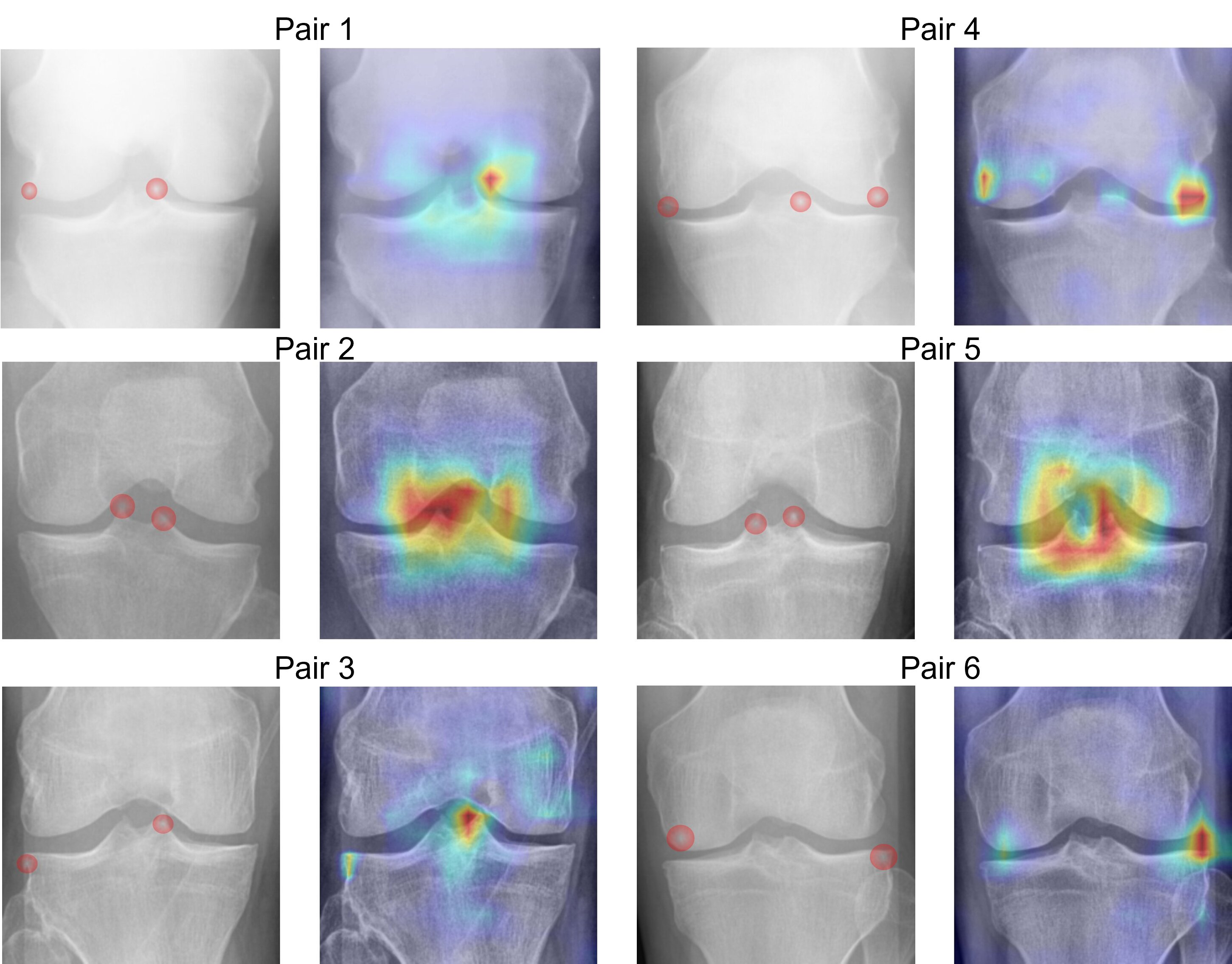}
\caption{Quantitative Consistency Analysis: Spatial overlap assessment between physician annotations and model GradCAM\cite{selvaraju2017gradcam} heatmaps demonstrated promising consistency, with the best-performing sample (Pair 5) achieving an IoU of 0.2852 and Dice coefficient of 0.4438. Pair 6 also exhibited strong performance (IoU: 0.2709, Dice: 0.4264, precision: 44.52\%). Notably, certain samples (Pair 3 and Pair 4) achieved 100\% recall, indicating complete coverage of physician-annotated critical pathological regions by the model. Overall, the model demonstrated robust recall capability across most samples (median: 51.37\%), effectively capturing diagnostically relevant features identified by clinical experts. The broader scope of model attention suggests potential identification of pathological information beyond manual annotations, providing quantitative support for interpretability research in deep learning-assisted pathological diagnosis.}
\label{fig:gradcam}
\end{figure*}

\begin{figure*}[!htbp]
\centering
\safeincludegraphics[width=0.70\textwidth]{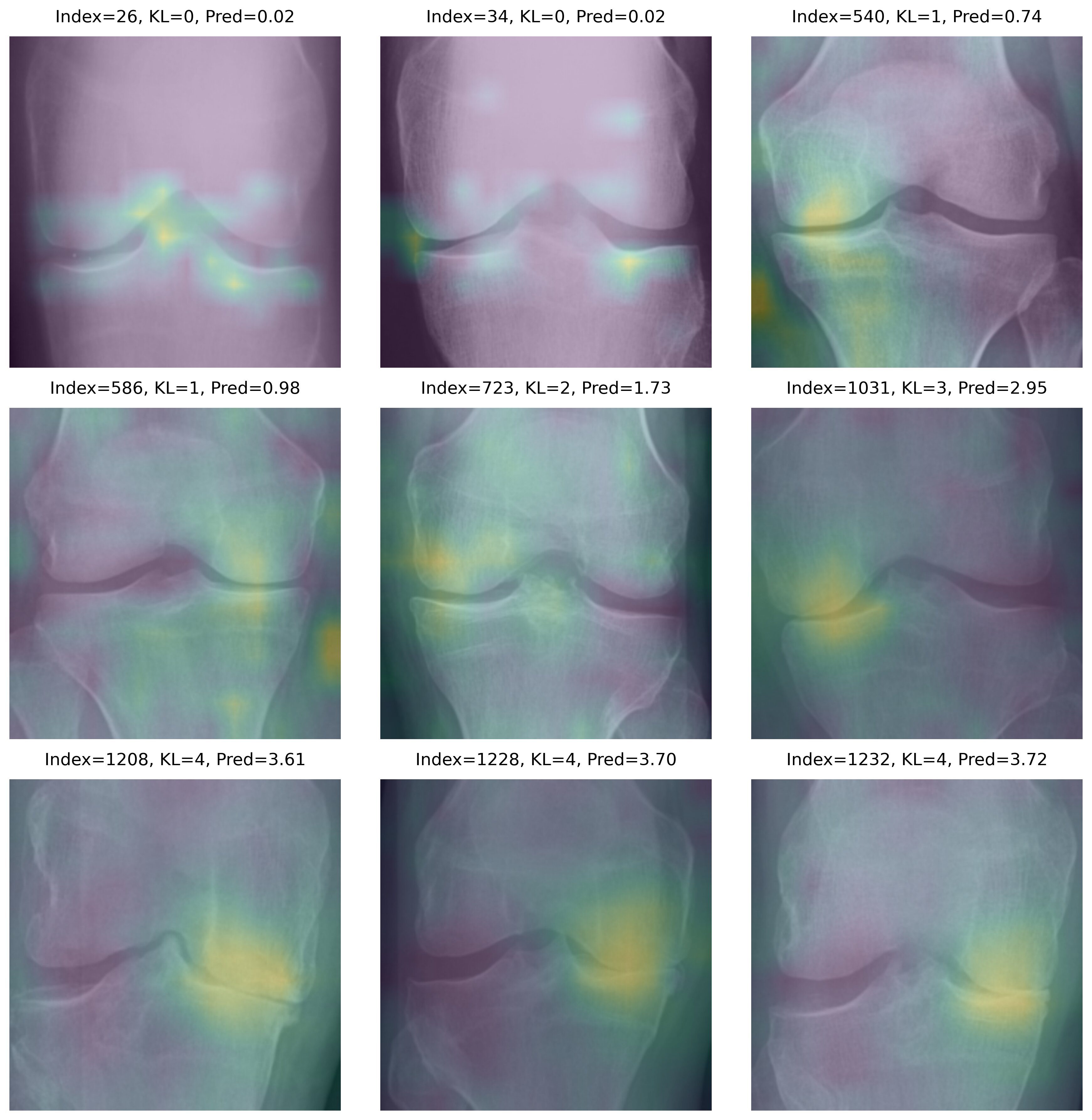}
\caption{\textbf{Evolution of Class Activation Mapping (CAM) with OA progression.}As OA exacerbates, the model's focal attention predominantly shifts and clusters at critical pathological sites, notably joint space narrowing (JSN). This shift illustrates that, empowered by a comprehensive global receptive field, the model autonomously scans for the most discriminative regions. Consequently, it precisely isolates localized pathological traits strongly associated with the predicted disease severity (Pred), reflecting a robust synthesis of macroscopic anatomical context and microscopic lesion characteristics.}
\label{fig:cam}
\end{figure*}

\subsection{Uncertainty quality}
\begin{figure*}[!htbp]
\centering
\safeincludegraphics[width=0.9\textwidth]{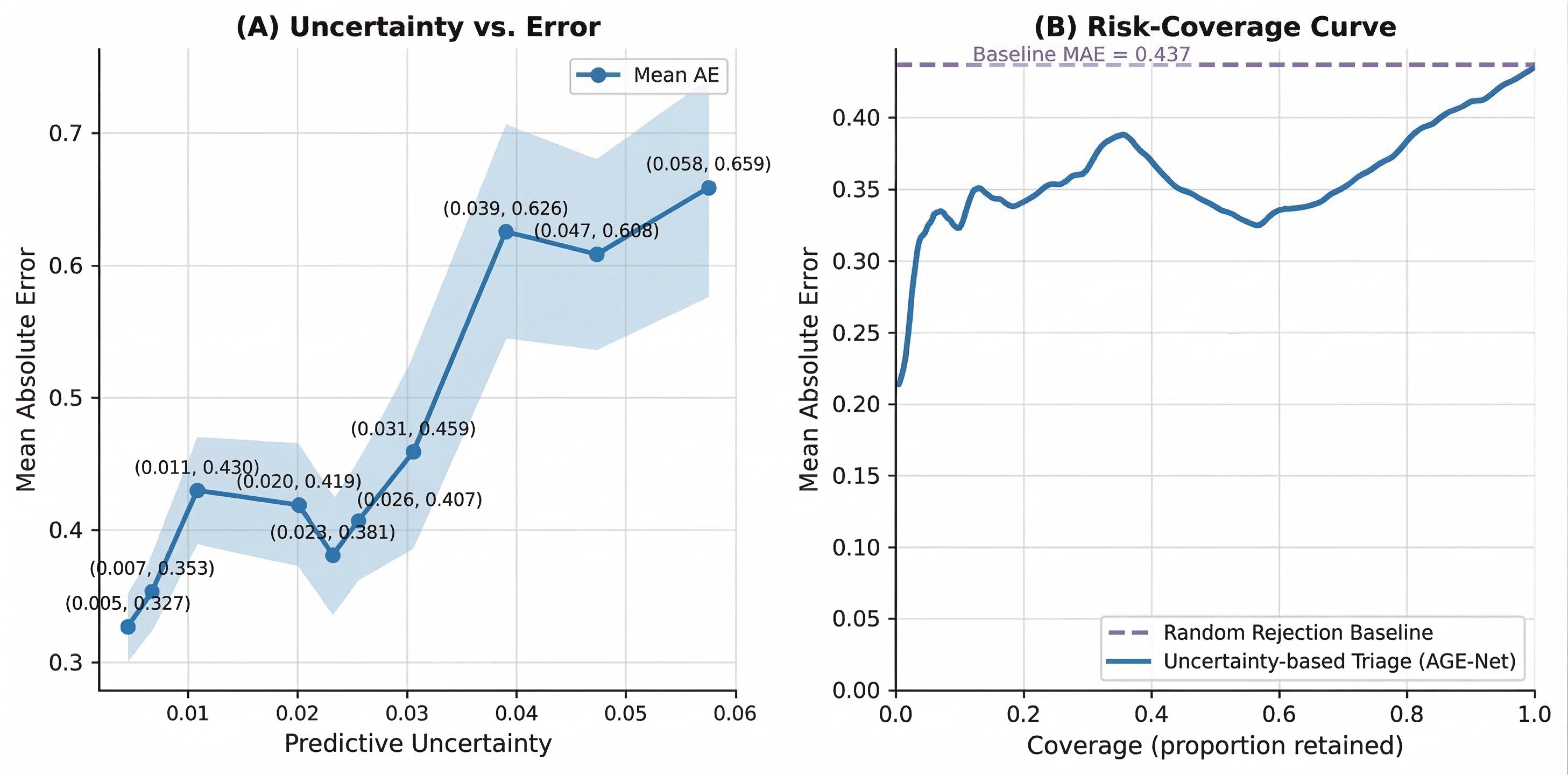}
\caption{Evaluation of uncertainty behavior. (Left) Mean AE (solid line) across uncertainty bins showing a strong positive correlation. The shaded region represents the 95\% confidence interval (CI). (Right) Risk--coverage curve obtained under uncertainty-based selective prediction, showing that excluding lower-confidence cases can improve retained predictive performance.}
\label{fig:uncertainty}
\end{figure*}

Figure \ref{fig:uncertainty} examines the relationship between the model's estimated uncertainty and prediction difficulty. In our experiments, higher uncertainty was generally associated with larger absolute error, suggesting that the evidential output captures aspects of predictive ambiguity. In addition, selective prediction based on uncertainty reduced empirical risk as coverage decreased. These findings support the potential utility of the uncertainty estimate for reliability-aware analysis, although more comprehensive calibration assessment would be needed before drawing stronger conclusions about deployment-oriented use.

\subsection{Robustness under perturbations}
\begin{figure*}[!htbp]
\centering
\safeincludegraphics[width=0.95\textwidth]{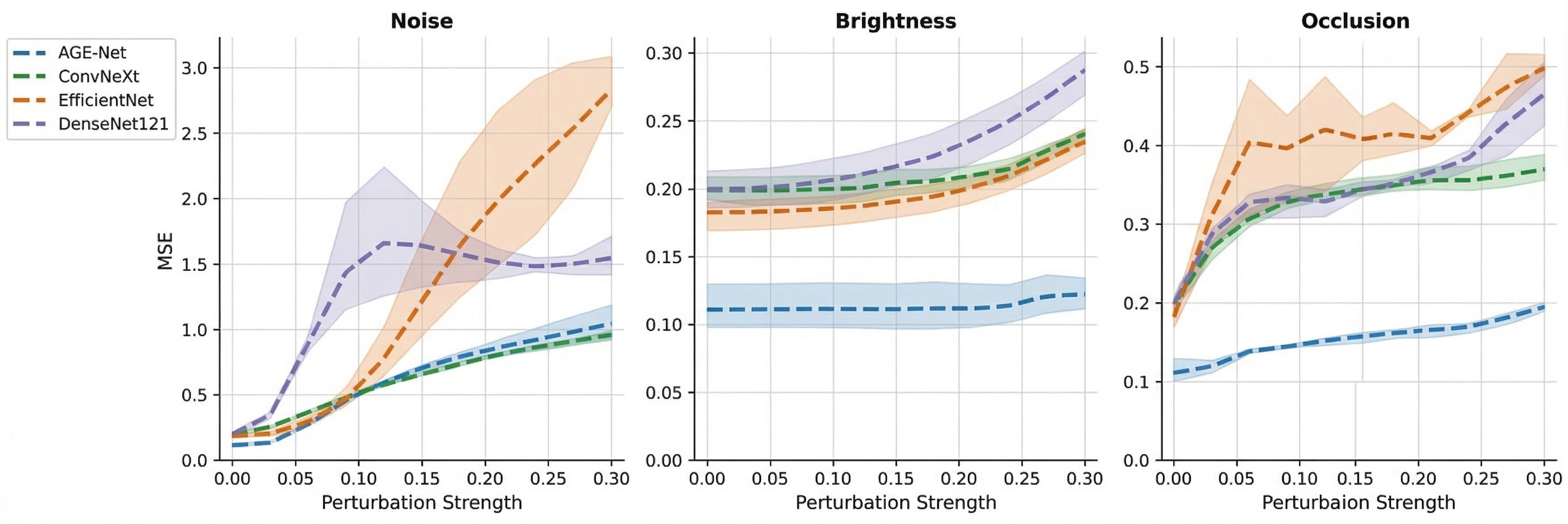}
\caption{\textbf{Extensive network robustness stress-testing under aggressive synthetic perturbations.} The graphical series plots computed Mean Squared Error (MSE) trajectories mapped strictly against linearly escalating perturbation intensities governing Gaussian noise injection, global photometric brightness shifts, and aggressive spatial field occlusion. AGE-Net maintains exceptional stability trajectories across the entire testing paradigm.}
\label{fig:robust}
\end{figure*}

Figure \ref{fig:robust} reports model performance under increasing levels of synthetic perturbation, including Gaussian noise, global brightness shifts, and spatial occlusion. Across these settings, AGE-Net maintained lower error and showed a slower degradation trend than the compared baseline models. This observation suggests that the proposed design provides a degree of robustness to common input perturbations within the evaluated experimental setup.

\section{Discussion}
\label{sec:discussion}
Automated KL grading sourced from raw, unedited knee radiographs represents a uniquely challenging computational frontier precisely because the clinically mandated diagnostic evidence is chronically subtle, physically dispersed across multiple anatomically distinct structures, and intrinsically fraught with deep subjectivity, especially regarding contiguous grade classification boundaries. The overarching architecture of AGE-Net was uniquely conceptualized to counter these exact morphological and operational characteristics. The sweeping performance advancements definitively documented within our experimental cohort perfectly parallel the synergistic integration of its custom modular components. The Spectral--Spatial Fusion (SSF) mechanism ensures the highly targeted, frequency-aware amplification of weak micro-textural and minute contrast variations while simultaneously enforcing stringent spatial selectivity through its proprietary gating parameter. This precision is absolutely invaluable for accurately classifying nascent, early-stage degenerative pathologies heavily obscured by substantial signal-to-noise degradations within the raw image data. Working in perfect tandem, the Anatomical Graph Reasoner (AGR) explicitly binds these microscopic findings into a cohesive whole, actively aggregating localized node information spanning vastly separated anatomical sectors utilizing highly dynamic message-passing protocols orchestrated across adaptive, latent-feature-derived neighborhoods. This process actively supports the non-local synthesis of inter-compartmental alignments and overarching structural geometry—critical diagnostic parameters functionally impossible to accurately resolve using rigidly constrained, purely localized convolutional operators. Concurrently, the Differential Refinement (DFR) module forcefully targets and amplifies high-gradient boundary transitions, deploying a mechanism structurally and philosophically aligned with the visual tracking of classic radiographic hallmarks such as joint-space narrowing and advancing bone contour erosion. Crucially, its deeply integrated residual architecture preserves fundamental underlying pixel representations and comprehensively prevents algorithmic overfitting to spurious, non-pathological edge artifacts. Ultimately, the synthesis of these carefully balanced design choices ensures that the framework flawlessly orchestrates multi-scale diagnostic evidence—seamlessly blending minute local textural appearances, sweeping global anatomical configurations, and hyper-focused boundary structures—in a manner fundamentally echoing the complex, multi-tiered visual synthesis naturally executed by highly trained musculoskeletal radiologists during active clinical review.

The foundational learning algorithm further solidifies the architecture's profound clinical viability. By decisively formulating the KL diagnostic spectrum not as a fragmented grouping of five fundamentally detached categorical bins, but as a rigorously intertwined, fluid ordinal hierarchy securely anchored by a dynamic pairwise ranking protocol, the network organically suppresses illogical, non-sequential monotonicity violations. This inherently blocks the system from sliding into computationally degenerate states capable of minimizing arbitrary pointwise error algorithms while simultaneously generating patently absurd, non-linear clinical progression curves. Simultaneously, the integration of the evidential COE-Head algorithm ensures the network deterministically outputs a highly robust severity mapping permanently coupled with an exceptionally calibrated mathematical uncertainty scalar. This specific capability unlocks immense potential regarding automated downstream clinical processing algorithms, allowing medical facilities to dynamically execute selective prediction and advanced triage protocols—such as instinctively deferring highly uncertain, diagnostically ambiguous image scans directly to secondary human specialist queues. When addressing translational deployment, several nuanced elements remain paramount. First, derived uncertainty variables must be constantly benchmarked against purely task-specific quality indicators, explicitly evaluating true calibration and dynamic risk-coverage trajectories to ensure the algorithm directly supports actionable diagnostic decisions rather than merely yielding abstract mathematical values. Second, overarching diagnostic stability must be incessantly validated via extensive external testing protocols completely spanning multiple, wholly independent imaging sites, variable hardware scanners, and fundamentally divergent patient populations, incorporating granular subgroup analysis tracking to preemptively isolate any latent algorithmic bias. Finally, given that foundational backbone selection profoundly impacts pretraining behaviors within specialized medical imaging domains, future benchmarking efforts against nascent, massively parameterized attention-based transformers must be strictly constrained within deeply controlled validation environments utilizing perfectly mirrored optimization configurations. Addressing these stringent elements will massively expand the evidence base supporting ordinal-consistent, uncertainty-aware KL grading systems deployed within high-volume, real-world radiological imaging environments.

\section{Conclusion}
\label{sec:conclusion}
In this work, we proposed AGE-Net, an anatomy-aware and ordinally constrained deep learning framework for automated KL grading of knee osteoarthritis from plain radiographs. By combining spectral--spatial enhancement, anatomical graph reasoning, pathology-aware differential refinement, and evidential ordinal prediction, the model achieved strong performance on both internal and external evaluations. The empirical results suggest that integrating anatomy-aware representation learning with ordinal supervision and uncertainty-related modeling can improve radiographic KOA assessment beyond a strong convolutional baseline. At the same time, the present findings should be interpreted within the scope of the evaluated datasets, preprocessing pipeline, and current calibration analysis. Further work is needed to strengthen statistical validation, assess uncertainty calibration more comprehensively, and examine generalizability across broader clinical populations and acquisition settings. Subject to such validation, AGE-Net may provide a useful framework for reliable computer-assisted KOA severity assessment.

\section*{CRediT authorship contribution statement}
\textbf{Xiaoyang Li}: Conceptualization, Methodology, Software, Investigation, Formal analysis, Visualization, Writing -- original draft, Writing -- review \& editing.
\textbf{Runni Zhou}: Methodology, Software, Investigation, Formal analysis, Writing -- original draft, Writing -- review \& editing.
\textbf{Xinghao Yan}: Data curation, Investigation, Writing -- review \& editing.
\textbf{Chenjie Zhu}: Data curation, Investigation, Writing -- review \& editing.
\textbf{Zhaochen Li}: Investigation, Validation, Writing -- review \& editing.
\textbf{Liehao Yan}: Investigation, Validation, Writing -- review \& editing.
\textbf{Rongrong Fu}: Resources, Data curation, Writing -- review \& editing.
\textbf{Yuan Chai}: Conceptualization, Supervision, Project administration, Funding acquisition, Writing -- review \& editing.

\section*{Declaration of competing interest}
The authors declare that they have no known competing financial interests or personal relationships that could have appeared to influence the work reported in this paper.

\section*{Acknowledgments}
The authors gratefully acknowledge Dr. Luxin Lou (Department of Radiology, Beijing Jishuitan Hospital, National Orthopaedic Medical Center) for providing the clinically validated region-of-interest annotations essential for the interpretability analysis presented in Fig. 10.

\appendix

\section{Additional implementation notes}
\label{sec:impl_notes}

\subsection{Training details and stability}
\label{sec:training_stability}
We leveraged highly optimized automatic mixed-precision protocols exclusively to massively augment base training throughput while substantially mitigating extreme GPU memory consumption variables. In practice, all forward and backward passes were executed under mixed-precision arithmetic with dynamic loss scaling to prevent numerical underflow, while all reported metrics were computed in full precision to ensure faithful evaluation. To fundamentally stabilize ongoing model evaluation metrics and radically compress arbitrary epoch-to-epoch performance variances, we continuously managed an Exponential Moving Average (EMA) encompassing all dynamic model parameters, subsequently generating all final reported metrics strictly utilizing these smoothed EMA weights during active test-time inference cycles. EMA effectively acts as a low-pass filter over optimization noise, yielding a more stable estimator of the underlying solution manifold, particularly under aggressive augmentation and long-horizon training schedules.

Given that the underlying infrastructural backbone was structurally initialized utilizing generic, externally pretrained ConvNeXt parameter weights, we proactively integrated a sophisticated layer-wise learning-rate decoupling algorithm. Specifically, the base backbone architecture was cautiously optimized utilizing a highly restricted learning rate to strictly prevent the catastrophic computational forgetting of inherently transferable base visual representations. Conversely, the completely novel integration modules and the complex predictive output head were accelerated using vastly larger learning rates to fiercely optimize incredibly granular, task-specific diagnostic mapping variables. Concretely, we applied a consistent multiplicative learning-rate ratio between backbone and newly introduced components throughout training, ensuring that the representation learned by the pretrained backbone remained stable while task-specific heads converged rapidly. Unless explicitly noted otherwise within specific documentation, all internal ablation permutations rigidly adhered to identical baseline optimization algorithms (including optimizer, schedule, augmentation, and EMA configuration) to absolutely guarantee strict comparative fairness across all metrics.

Furthermore, empirical testing identified a deeply critical operational intersection spanning aggressive data augmentation deployments and underlying ordinal constraint algorithms. Notably, we completely deactivated the overarching ranking loss protocol explicitly whenever dynamic Mixup augmentation was activated. Because Mixup naturally synthesizes interpolated, highly softened categorical labels, it immediately renders the strict discrete inequality constraint $y_i - y_j \ge 1$ mathematically ill-defined, thereby threatening to inject structurally inconsistent supervision noise directly into the ordinal training arrays. Formally, when Mixup produces $\tilde{y}=\lambda y_i + (1-\lambda) y_j$ with $\lambda\in(0,1)$, the pairwise ordering relation ceases to be well-posed under the discrete KL hierarchy, and na\"ively enforcing ranking constraints can bias the continuous severity regressor toward arbitrary margin violations. Under these specific augmented instances, the network architecture was meticulously optimized relying strictly upon the primary evidential regression loss algorithms, thereby preserving a coherent optimization objective while still benefiting from Mixup-induced regularization.

\subsection{Test-time augmentation}
\label{sec:tta}
During formal inference evaluation passes, we operationalized a lightweight, highly targeted horizontal flip-based Test-Time Augmentation (TTA) strategy. This algorithm mathematically synthesized dual predictive probabilities derived equally from the pristine, original radiographic image and its perfectly mirrored horizontal counterpart. This specialized augmentation routine drastically suppresses inherent predictive variance while massively increasing overall network resilience to incredibly minor patient pose deviations and isolated hardware acquisition artifacts, all completely without expanding the baseline training complexity curve.

For the evidential head, we applied aggregation directly at the level of predicted distributional parameters to preserve internal consistency. Concerning the specialized evidential head protocols, we precisely aggregated and averaged the dual predicted severity scores $\gamma$, strictly utilizing the resultant unified average to drive the final, discrete grade rounding calculations. All associated underlying mathematical uncertainty parameters were identically managed utilizing strictly congruent score aggregation methodologies completely consistent with our baseline implementation parameters, ensuring that the final uncertainty estimate reflects consensus evidence across both views rather than a single-pass artifact. This simple two-view TTA was selected to balance robustness gains and computational cost, and was applied uniformly to all methods for equitable comparison wherever applicable.

\section{Extended ablations and statistical analysis}
\label{sec:planned}

\subsection{Planned extended ablations}
To comprehensively dissect the exact, isolated mathematical contribution of every singular architectural parameter and to thoroughly validate the overarching stability of all pre-selected hyperparameter settings, we have thoroughly engineered the following subsequent extended ablation protocols. All ablations will be executed under an identical training recipe (data splits, augmentation suite, optimizer, schedule, and evaluation pipeline), and will be repeated across multiple random seeds to reduce variance and enable statistically meaningful paired comparisons:
\begin{itemize}
  \item \textbf{w/o SSF:} The absolute excision of all frequency modulation and spatial gating nodes explicitly to isolate and quantify the absolute baseline contribution provided exclusively by the spectral--spatial algorithmic enhancement arrays. Beyond aggregate metrics, we will additionally inspect whether the removal disproportionately degrades performance on early-grade cases, where high-frequency micro-textures are expected to be most informative.
  \item \textbf{w/o DFR:} The total removal of the differential refinement protocols definitively isolating the intrinsic value of gradient-based, boundary-sensitive structural feature enhancements, with targeted analysis on borderline grades where joint-space narrowing and marginal contour cues dominate the diagnostic decision.
  \item \textbf{w/o COE:} The complete substitution of the advanced evidential NIG head with a vastly simplified, canonical heteroscedastic regression parameter, exclusively to benchmark and map complex, comparative predictive uncertainty behaviors across diverse boundaries. In this setting, we will evaluate both accuracy and the calibration behavior of uncertainty proxies under identical inference rules.
  \item \textbf{$k$ sensitivity:} Systematically modulating the discrete $k$ value within the active AGR module (testing subsets including 5/9/13) to rigorously plot the precise architectural effect dynamic neighborhood size generates regarding complex non-local reasoning fidelity, and to identify the stability range that best balances anatomical context capture against over-smoothing.
  \item \textbf{Token grid size:} Drastically varying the foundational localized pooling grid spatial resolution arrays (evaluating subsets such as $10\times10$, $14\times14$, $16\times16$) to scientifically benchmark the precise computational trade-off separating localized token granularity against sweeping processing hardware costs, including potential impacts on interpretability localization granularity and runtime latency.
\end{itemize}

\subsection{Planned statistical analysis}
To quantify the statistical reliability of all reported improvements, we will perform paired comparisons across matched random seeds when benchmarking the intact AGE-Net against major ablations and the bare ConvNeXt baseline. Specifically, primary metrics (e.g., QWK and MSE) will be evaluated using paired tests on per-seed summaries (e.g., paired $t$-tests) and accompanied by effect sizes. We will additionally report confidence intervals via nonparametric bootstrap resampling on the test set (patient-level resampling where applicable) to account for correlation structure and provide distribution-free uncertainty estimates. When testing multiple ablation variants, we will apply multiple-comparison control to mitigate inflated false discoveries, and emphasize effect sizes alongside $p$-values to reflect practical relevance.

\section{Additional confusion matrices}
\label{sec:conf_mats_app}
\begin{figure*}[!htbp]
\centering
\safeincludegraphics[width=\textwidth]{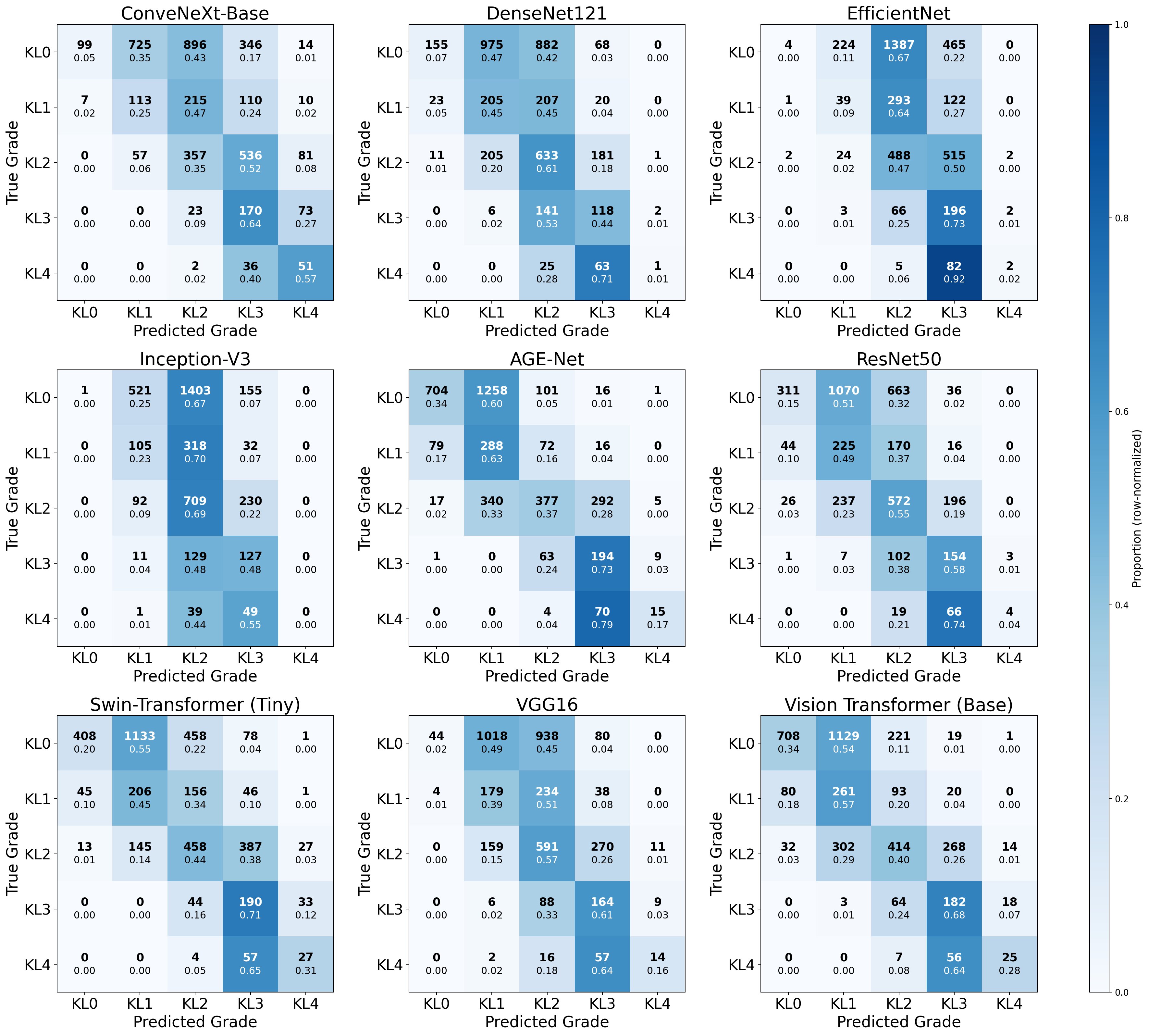}
\caption{Row-normalized confusion matrices for AGE-Net and representative baseline models. AGE-Net shows a larger proportion of errors concentrated near adjacent grades, whereas several baseline models exhibit a broader distribution of off-diagonal errors. These matrices complement aggregate metrics by illustrating the error structure along the KL severity continuum.}
\label{fig:confusion_combined}
\end{figure*}

Figure \ref{fig:confusion_combined} exhaustively cross-references and evaluates complex row-normalized confusion matrix distributions dynamically mapping the completely intact AGE-Net framework specifically against prominent baseline convolutional architectures. In addition to the qualitative structure of off-diagonal mass, these matrices facilitate a more granular inspection of error modes, including whether misclassifications concentrate within adjacent-grade boundaries and whether any method exhibits clinically implausible long-range grade inversions. Such per-grade diagnostics provide complementary evidence to aggregate metrics, clarifying where architectural changes yield practical gains along the KL severity continuum.

\ModuleFloatBarrier
\clearpage

\bibliographystyle{cas-model2-names}
\bibliography{refsV3}

\end{document}